%% file: main.tex
\documentclass[10pt,twocolumn]{article}

\usepackage[margin=0.72in,columnsep=0.25in]{geometry}
\usepackage[T1]{fontenc}
\usepackage[utf8]{inputenc}
\usepackage{lmodern}
\usepackage{microtype}
\usepackage{xspace}
\usepackage{amsmath,amssymb,mathtools,amsthm}
\usepackage{booktabs,tabularx,multirow,makecell,array}
\usepackage{graphicx}
\usepackage{xcolor}
\usepackage{enumitem}
\usepackage{algorithm}
\usepackage[noend]{algpseudocode}
\usepackage{tikz}
\usetikzlibrary{arrows.meta,positioning,fit,calc,backgrounds,shapes.geometric,decorations.pathreplacing}
\usepackage{pgfplots}
\usepackage{pgfplotstable}
\pgfplotsset{compat=1.18}
\usepackage{subcaption}
\usepackage{url}
\usepackage[round,authoryear]{natbib}
\usepackage[colorlinks=true,allcolors=blue!55!black]{hyperref}
\usepackage[nameinlink,capitalize,noabbrev]{cleveref}
\usepackage{balance}

\input{metadata}
\input{macros}

\newtheorem{proposition}{Proposition}
\newtheorem{definition}{Definition}

\setlist[itemize]{leftmargin=*,topsep=2pt,itemsep=1pt}
\setlist[enumerate]{leftmargin=*,topsep=2pt,itemsep=1pt}
\title{\papertitle}
\author{\paperauthors}
\date{\paperdate}

\begin{document}
\maketitle

\begin{abstract}
Credit assignment in large-language-model reinforcement learning (LLM RL) can be separated into three objects: \emph{evidence} about success, a \emph{transport operator} that converts this evidence into token-level advantages, and an \emph{update geometry} that turns advantages into policy changes. Recent work has greatly improved evidence, sampling, and update geometry, but the transport operator is usually architecture-agnostic. Fixed-discount GAE applies a stationary geometric kernel along token time; group-relative methods broadcast an outcome statistic across an entire response. Neither operator represents the trajectory-specific computation used by the Transformer policy itself.

We introduce \emph{computation-conditioned credit transport} (\framework), a general framework in which a detached statistic of the behavior policy's internal computation parameterizes the causal kernel that transports downstream value through a rollout. Our concrete algorithm, \method, maps native attention concentration to a bounded per-token retention gate, uses the gate in both the one-step bootstrap and a path-dependent generalized-advantage trace (\compgae), and co-designs a transport-aligned critic (\tac) that reuses the actor's hidden states and routing information without a second same-scale Transformer. The task reward and clipped PPO policy objective remain unchanged; a constant gate recovers fixed-coefficient GAE.

Across five Qwen3-4B seeds, \method reaches $61.4\%$ final held-out development accuracy (95\% CI $[60.8,62.0]$), versus $53.8\%$ $[52.9,54.7]$ for a separately tuned GRPO baseline. A controlled $2\!\times\!2$ gate-by-critic experiment shows that neither \compgae with a standard critic ($55.2\%$) nor \tac with a mean-matched fixed gate ($56.4\%$) reproduces the full result; the seed-matched final interaction is $+2.4$ points $[1.9,2.9]$. Gate shuffling and position-only controls show that trajectory-specific alignment matters beyond a favorable marginal schedule. In a matched PPO stress grid, \method is stable in $10/12$ runs versus $3/12$ for PPO. Final-checkpoint frozen evaluation improves over GRPO by $4.3$ and $3.9$ greedy pass@1 macro points on Qwen3-4B and Llama-3.1-8B-Instruct, respectively. These results establish policy-internal computation as a useful estimator variable and open a broader research program on architecture-aware credit kernels, traces, and critics.
\end{abstract}


\input{sections/01_introduction}
\input{sections/02_credit_coordinates}
\input{sections/03_framework}
\input{sections/04_method}
\input{sections/05_theory}
\input{sections/06_experiments}
\input{sections/07_results}
\input{sections/08_related_work}
\input{sections/09_discussion}
\input{sections/10_conclusion}

\balance
\bibliographystyle{plainnat}
\bibliography{references}

\clearpage
\appendix
\onecolumn
\input{sections/A_algorithm}
\input{sections/B_full_results}
\input{sections/C_reproducibility}
\input{sections/D_proofs}
\input{sections/E_limitations}

\end{document}

%% file: metadata.tex
\newcommand{\papertitle}{Let Credit Follow Computation: Architecture-Aware Credit Transport for Large Language Model Reinforcement Learning}

\newcommand{\paperauthors}{Qifan Shi \and Zhaolu Kang \and Chenghua Zhu}

\newcommand{\paperdate}{August 2026}

%% file: macros.tex
\newcommand{\method}{\textsc{CompPO}\xspace}
\newcommand{\framework}{\textsc{CCT}\xspace}
\newcommand{\oldmethod}{\textsc{NMPO}\xspace}
\newcommand{\compgae}{\textsc{Comp-GAE}\xspace}
\newcommand{\tac}{\textsc{TAC}\xspace}
\newcommand{\E}{\mathbb{E}}

\newcommand{\ind}{\mathbf{1}}
\newcommand{\kappamin}{\kappa_{\mathrm{lo}}}
\newcommand{\kappamax}{\kappa_{\mathrm{hi}}}

\DeclareMathOperator{\clip}{clip}

\DeclareMathOperator{\softmax}{softmax}

\DeclareMathOperator{\diag}{diag}

%% file: sections/01_introduction.tex
\section{Introduction}
\label{sec:intro}

Reinforcement learning with verifiable or preference-derived rewards has become a primary route to stronger large-language-model (LLM) reasoning and decision making. The algorithmic landscape has advanced quickly: PPO supplies bootstrapped token-level advantages \citep{schulman2017ppo,schulman2015gae}; RLOO and GRPO replace the critic with response-level comparison \citep{ahmadian2024rloo,shao2024deepseekmath}; DAPO, Dr.~GRPO, and GSPO improve clipping, normalization, sampling, and importance-ratio geometry \citep{yu2025dapo,liu2025drgrpo,zheng2025gspo}; process-reward and rollout-based methods generate denser evidence about intermediate decisions \citep{lightman2023prm,uesato2022process,cui2025prime,kazemnejad2024vineppo}. Yet one part of the learning stack remains comparatively implicit: \emph{the operator that transports delayed evidence through the generated trajectory}.

This omission matters because a Transformer policy is not a homogeneous token chain. At each position it performs a content-dependent computation over the prefix: heads retrieve different events, representations compose them, and the active dependency pattern changes across trajectories. A late mathematical conclusion may use an early substitution; a tool action may depend on an observation many turns earlier; a code token may be constrained by a distant declaration. The policy is therefore architecture-rich and history-structured. Its credit estimator is usually not. Fixed-discount GAE uses one continuation coefficient at every transition; group-relative methods assign a response-level statistic to all valid tokens in the response. We call this discrepancy the \textbf{architecture--credit mismatch}.

To make the mismatch precise, we factor a token-level policy-gradient pipeline into three conceptually distinct objects:
\begin{enumerate}
    \item \textbf{credit evidence}: what indicates that a decision sequence was good or bad---terminal verifiers, process rewards, judges, counterfactual continuations, or critic residuals;
    \item \textbf{credit transport}: the causal operator that moves that evidence across positions or states to produce token-level advantages; and
    \item \textbf{update geometry}: importance ratios, clipping, KL constraints, entropy terms, and normalization that convert advantages into a policy update.
\end{enumerate}
This factorization clarifies why many powerful LLM-RL advances are complementary rather than interchangeable. Better verifiers change evidence. GRPO changes the baseline and sampling statistic. GSPO changes update geometry. A process reward changes where evidence enters. None of these choices uniquely determines how downstream evidence should be transported through the policy's own computation.

For standard GAE, the transport can be written as a stationary upper-triangular kernel. If $\boldsymbol\delta$ is the vector of TD residuals, then
\begin{equation}
    \mathbf A = \mathbf K^{\gamma,\lambda}\boldsymbol\delta,
    \qquad
    K^{\gamma,\lambda}_{t,u}
    =\ind[u\ge t](\lambda\gamma)^{u-t}.
    \label{eq:intro_gae_kernel}
\end{equation}
The kernel depends only on temporal separation. A canonical group-relative estimator instead produces $\mathbf A^{(i)}=b_i\mathbf 1$ within response $i$: a rank-one broadcast geometry with no within-response transport structure. Both are coherent estimators; both discard the realized internal computation of the policy.

Our thesis is:
\begin{quote}
\emph{The policy already computes a trajectory-specific dependency structure; credit transport should be allowed to condition on that computation.}
\end{quote}
We introduce \textbf{computation-conditioned credit transport} (\framework). Let $\Phi_{\theta^-}(h_t)$ be a statistic extracted from the behavior policy's internal computation at prefix $h_t$, and let a bounded map produce a retention gate $\kappa_t$. The resulting path kernel is
\begin{equation}
K^{\kappa,\lambda}_{t,u}
=\ind[u\ge t]\lambda^{u-t}\prod_{j=t}^{u-1}\kappa_j.
\label{eq:intro_cct_kernel}
\end{equation}
Unlike \cref{eq:intro_gae_kernel}, it is nonstationary and trajectory-specific: the sensitivity of $A_t$ to a later residual $\delta_u$ is the realized product of gates along the intervening computation. The gate is computed by the old policy, detached, and stored with the rollout. Thus \framework changes the estimator-side transport kernel, not the environment reward or task objective.

We instantiate the framework as \textbf{Computation-Conditioned Policy Optimization} (\method).\footnote{The anonymous submission used the working name \oldmethod and denoted the gate by $\alpha_t$. We rename the framework and algorithm to foreground the more general contribution and to avoid conflating estimator-side retention with task discounting. The implemented mechanism and objective experimental results are unchanged.} \method uses native attention concentration as an inexpensive scalar summary of whether the current computation is dominated by a sparse subset of history or supported diffusely. The gate enters a one-step computation-conditioned bootstrap and a path-dependent generalized-advantage trace, \compgae. A transport-aligned critic (\tac) pools value-relevant history from the actor's existing hidden states and attention and uses the same gate to couple local and routed-history value estimates. The outer clipped PPO objective, verifier, rollout format, and reward are unchanged.

The method is deliberately more than an attention-weighted loss. Attention does not directly multiply the policy gradient. It parameterizes the bootstrap coefficient, every downstream residual's trace weight, and the critic state geometry used to estimate those targets. Nor do we claim that attention is causal token attribution. Concentration is a policy-native structural proxy, and its usefulness must be established through controlled alignment tests.

The empirical evidence is organized to test the main alternative explanations directly. First, GRPO and \method are screened over the same $3\!\times\!3$ actor-learning-rate/KL grid and confirmed with five seeds. \method reaches $61.4\%$ final held-out development accuracy, compared with $53.8\%$ for GRPO, and every \method seed exceeds every GRPO seed at both best and final endpoints. Second, a five-seed $2\!\times\!2$ experiment crosses dynamic versus mean-matched fixed transport with standard versus aligned critics. Dynamic transport with the standard critic reaches $55.2\%$ final accuracy; the aligned critic with a fixed gate reaches $56.4\%$; full \method reaches $61.4\%$, with a positive final interaction of $+2.4$ points $[1.9,2.9]$. The standard critic has negative explained variance and weak TD-target rank correlation, whereas the aligned critic tracks the induced targets substantially better. Third, global gate shuffling and a frozen position-only schedule reduce performance, showing that the result is not explained by marginal coefficient scale or position alone. Fourth, a matched PPO stress grid supports a bounded robustness claim: \method is stable in $10/12$ runs versus $3/12$ for PPO. Finally, final-checkpoint frozen evaluation improves over GRPO by $4.3$ and $3.9$ greedy macro points on Qwen3-4B and Llama-3.1-8B-Instruct.

The deeper implication is that LLM RL need not remain architecture-agnostic. The dominant critic-free trend may reflect a contingent mismatch between conventional value representations and long-horizon targets, rather than an intrinsic impossibility of actor--critic learning for language models. More broadly, attention, retrieval, memory access, MoE routing, tool dependencies, or cross-modal information flow could parameterize future Bellman-style operators, traces, critics, and trust regions.

Our contributions are:
\begin{itemize}
    \item \textbf{A factorization and problem abstraction.} We separate credit evidence, credit transport, and update geometry, identify the architecture--credit mismatch, and characterize fixed-discount GAE and group-relative broadcast as two architecture-agnostic transport geometries.
    \item \textbf{A general framework.} \framework turns detached policy-internal computation into a path-dependent causal transport kernel. We derive its matrix form, sensitivity interpretation, reduction to fixed GAE, contraction, and bounded-trace properties.
    \item \textbf{A concrete algorithm.} \method couples an attention-derived retention gate, \compgae, and a transport-aligned actor-feature critic while preserving the external reward and clipped PPO update.
    \item \textbf{Mechanism-level evidence.} Matched tuning, five-seed factorials, critic diagnostics, shuffling, position-only schedules, a PPO stress grid, a stronger group baseline, and two-backbone frozen evaluation isolate what the mechanism does and where the evidence stops.
    \item \textbf{A research agenda.} The work elevates policy-internal computation to a first-class RL estimator variable and motivates architecture-aware credit transport beyond the scalar attention instantiation studied here.
\end{itemize}

%% file: sections/02_credit_coordinates.tex
\section{Credit Assignment as Evidence, Transport, and Update}
\label{sec:coordinates}

\input{figures/credit_coordinates}

\subsection{Trajectory setup}
Let $x$ be a prompt and $y_{1:T}$ a response sampled from $\pi_\theta(y_t\mid h_t)$, where $h_t=(x,y_{<t})$. Let $r_t$ denote externally specified reward evidence at position $t$; the main experiments use a sparse terminal verifier. A policy-gradient method ultimately constructs token coefficients $A_t$ and applies them through an update rule such as PPO. We write this process abstractly as
\begin{equation}
\begin{aligned}
    \underbrace{\mathbf e}_{\text{evidence}}
    &\xrightarrow{\;\mathbf K\;}
    \underbrace{\mathbf A}_{\text{transported credit}}
    \xrightarrow{\;\mathcal U\;}
    \underbrace{\Delta\theta}_{\text{policy update}},\\[-1mm]
    &\hspace{18mm}\mathbf A=\mathbf K\mathbf e.
\end{aligned}
\label{eq:factorization}
\end{equation}
Here $\mathbf e$ may contain rewards, TD residuals, process scores, counterfactual estimates, or other local evidence. $\mathbf K$ is a causal transport operator, and $\mathcal U$ contains likelihood ratios, clipping, KL, entropy, and normalization. The factorization is analytic rather than exclusive: an algorithm may jointly modify all three terms. Its purpose is to identify which object a contribution changes.

\subsection{Fixed-discount GAE is stationary temporal transport}
With a value function $V_\psi$, standard GAE uses
\begin{align}
\delta_t^\gamma &= r_t+\gamma V_\psi(h_{t+1})m_{t+1}-V_\psi(h_t)m_t,\\
A_t^{\gamma,\lambda}&=\delta_t^\gamma+\gamma\lambda A_{t+1}^{\gamma,\lambda}m_{t+1},
\label{eq:stdgae}
\end{align}
where $m_t$ masks invalid or padded positions. Ignoring masks for notation and stacking residuals gives
\begin{equation}
    \mathbf A=(\mathbf I-\lambda\gamma\mathbf S)^{-1}\boldsymbol\delta,
    \label{eq:std_matrix}
\end{equation}
where $(\mathbf S\mathbf v)_t=v_{t+1}$. Therefore
\begin{equation}
K^{\gamma,\lambda}_{t,u}
=\ind[u\ge t](\lambda\gamma)^{u-t}.
\label{eq:toeplitz}
\end{equation}
The transport kernel is stationary and Toeplitz: it depends only on token distance. This is not equivalent to saying that GAE has no local signal; TD bootstrapping can make $\delta_t$ informative even with sparse reward. The structural limitation is narrower: the continuation geometry contains no variable representing how the policy computed the realized action.

\subsection{Group-relative credit is response-level broadcast}
For $n$ responses to the same prompt, a canonical group-relative baseline is
\begin{equation}
 b_i=\frac{R_i-\operatorname{mean}_j R_j}{\operatorname{std}_j R_j+\varepsilon},
 \qquad
 \mathbf A^{(i)}=b_i\mathbf 1_{T_i}.
 \label{eq:group_broadcast}
\end{equation}
Within a response, this is a rank-one broadcast geometry: every valid token receives the same outcome-derived coefficient. DAPO, Dr.~GRPO, and GSPO materially improve clipping, normalization, sampling, or sequence-ratio behavior \citep{yu2025dapo,liu2025drgrpo,zheng2025gspo}, but do not by themselves insert a trajectory-specific Transformer-computation variable into a Bellman/GAE transport kernel.

When a sampled group has identical rewards, the standardized group evidence may vanish; when rewards differ, the estimator still does not distinguish tokens that were pivotal from routine tokens. These properties help explain why fine-grained credit has become a major LLM-RL research direction \citep{zhang2026creditsurvey}.

\subsection{Architecture--credit mismatch}
A causal Transformer generates each action through a realized internal computation graph. Two responses can have identical lengths, rewards, and group statistics yet route information through different historical tokens, heads, memories, or experts. An estimator that assigns them the same transport kernel discards this difference.

\begin{definition}[Architecture--credit mismatch]
A credit estimator has an architecture--credit mismatch when its transport operator is invariant to changes in the policy's realized internal computation that may alter which history is used to produce the optimized actions.
\end{definition}

This definition does not imply that internal computation is a faithful causal explanation. It identifies an unused information channel. Whether that channel improves learning is an empirical question requiring controls for critic capacity, position, marginal gate distribution, and training engineering.

\subsection{Three coordinates, not three mutually exclusive algorithms}
\Cref{fig:coordinates} visualizes temporal, group/outcome, and computational coordinates. They are not mutually exclusive sources of evidence. A process reward can be transported temporally or computationally; a group baseline can be combined with a token trace; a sequence-level importance ratio can consume an architecture-aware advantage. The contribution of this paper is to make the transport coordinate explicit and optimizable rather than treating fixed time or response broadcast as unavoidable defaults.

%% file: figures/credit_coordinates.tex
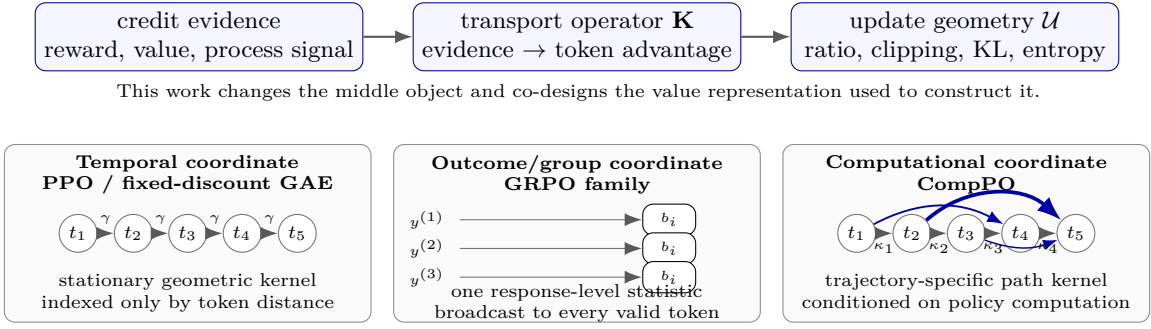
\begin{figure*}[t]
\centering
\begin{tikzpicture}[font=\small,>=Latex,
    token/.style={circle,draw=black!55,fill=white,minimum size=5.0mm,inner sep=0pt,font=\scriptsize},
    box/.style={rounded corners,draw=black!45,fill=black!2,minimum height=2.35cm,minimum width=4.85cm,inner sep=3pt},
    stage/.style={rounded corners,draw=blue!55!black,fill=blue!4,minimum height=8mm,text width=4.1cm,align=center},
    ttl/.style={font=\bfseries\scriptsize,align=center,text width=4.55cm},
    desc/.style={font=\scriptsize,align=center,text width=4.45cm},
    arr/.style={->,thick,draw=black!65}]

\node[stage] (e) at (-5.0,1.85) {credit evidence\\reward, value, process signal};
\node[stage] (k) at (0,1.85) {transport operator $\mathbf K$\\evidence $\rightarrow$ token advantage};
\node[stage] (u) at (5.0,1.85) {update geometry $\mathcal U$\\ratio, clipping, KL, entropy};
\draw[arr] (e) -- (k);
\draw[arr] (k) -- (u);
\node[font=\scriptsize,align=center,text width=12.5cm] at (0,1.12) {This work changes the middle object and co-designs the value representation used to construct it.};

\node[box] (a) at (-5.15,-0.75) {};
\node[ttl] at ($(a.center)+(0,0.78)$) {Temporal coordinate\\PPO / fixed-discount GAE};
\foreach \x/\i in {-1.45/1,-0.72/2,0/3,0.72/4,1.45/5}{\node[token] (a\i) at ($(a.center)+(\x,0.0)$) {$t_{\i}$};}
\foreach \i/\j in {1/2,2/3,3/4,4/5}{\draw[arr] (a\i) -- node[above,font=\tiny] {$\gamma$} (a\j);}
\node[desc] at ($(a.center)+(0,-0.78)$) {stationary geometric kernel\\indexed only by token distance};

\node[box] (b) at (0,-0.75) {};
\node[ttl] at ($(b.center)+(0,0.78)$) {Outcome/group coordinate\\GRPO family};
\foreach \y/\i in {0.18/1,-0.20/2,-0.58/3}{
  \node[font=\tiny,anchor=east] at ($(b.center)+(-1.63,\y)$) {$y^{(\i)}$};
  \draw[black!55] ($(b.center)+(-1.55,\y)$) -- ($(b.center)+(0.75,\y)$);
  \node[draw,rounded corners,fill=white,font=\tiny,minimum width=7mm,minimum height=4mm] (r\i) at ($(b.center)+(1.22,\y)$) {$b_i$};
  \draw[arr] ($(b.center)+(0.75,\y)$) -- (r\i);
}
\node[desc] at ($(b.center)+(0,-0.92)$) {one response-level statistic\\broadcast to every valid token};

\node[box] (c) at (5.15,-0.75) {};
\node[ttl] at ($(c.center)+(0,0.78)$) {Computational coordinate\\\method};
\foreach \x/\i in {-1.45/1,-0.72/2,0/3,0.72/4,1.45/5}{\node[token] (c\i) at ($(c.center)+(\x,0.0)$) {$t_{\i}$};}
\draw[->,line width=0.7pt,draw=blue!65!black,bend left=28] (c1) to (c4);
\draw[->,line width=1.4pt,draw=blue!65!black,bend left=42] (c2) to (c5);
\draw[->,line width=0.55pt,draw=blue!65!black,bend right=20] (c3) to (c5);
\foreach \i/\j/\kg in {1/2/{\kappa_1},2/3/{\kappa_2},3/4/{\kappa_3},4/5/{\kappa_4}}{
  \draw[arr] (c\i) -- node[below,font=\tiny] {$\kg$} (c\j);
}
\node[desc] at ($(c.center)+(0,-0.78)$) {trajectory-specific path kernel\\conditioned on policy computation};

\end{tikzpicture}
\caption{Credit assignment factorized into evidence, transport, and update geometry (top), with three transport coordinates used in LLM RL (bottom). Fixed GAE uses a stationary temporal kernel; group-relative methods broadcast response-level evidence; \method lets behavior-policy computation parameterize the causal transport path.}
\label{fig:coordinates}
\end{figure*}

%% file: sections/03_framework.tex
\section{Computation-Conditioned Credit Transport}
\label{sec:framework}

\subsection{Policy-internal computation as an estimator variable}
Let $\theta^-$ denote the behavior policy that generated the rollout. Its forward pass exposes a computation object $\mathcal G_{\theta^-}(h_t)$, such as attention, retrieval scores, memory access, expert routing, or cross-modal flow. A statistic $\Phi$ and bounded map $g$ produce
\begin{equation}
    z_t=\Phi(\mathcal G_{\theta^-}(h_t)),
    \qquad
    \kappa_t=g(z_t),
    \qquad 0\le \kappa_t\le \bar\kappa<1.
    \label{eq:kappa_general}
\end{equation}
The gate is stored with the rollout and detached. The same realized $\kappa_t$ is used throughout the optimization epochs for that batch. This behavior-policy conditioning is central: \framework does not differentiate through a policy-dependent task discount and does not redefine the environment's utility. It constructs a rollout-conditioned credit estimator.

\subsection{Computation-conditioned bootstrap and trace}
For a value function over a computation-aware state representation $s_t^{\mathrm C}$, define
\begin{align}
\delta_t^\kappa
&=r_t+\kappa_t V_\psi(s_{t+1}^{\mathrm C})m_{t+1}
      -V_\psi(s_t^{\mathrm C})m_t,
\label{eq:comp_td}\\
A_t^{\kappa,\lambda}
&=\delta_t^\kappa+\lambda\kappa_t A_{t+1}^{\kappa,\lambda}m_{t+1}.
\label{eq:comp_gae}
\end{align}
We call \cref{eq:comp_gae} \compgae. The gate has two linked effects: it changes the one-step bootstrap target and the retention of all later residuals across the current transition.

With $\mathbf D_\kappa=\diag(\kappa_1,\ldots,\kappa_T)$ and shift matrix $\mathbf S$, the valid-token recursion is
\begin{equation}
\begin{aligned}
    \mathbf A&=\boldsymbol\delta+\lambda\mathbf D_\kappa\mathbf S\mathbf A,\\
    \mathbf A&=\mathbf K^{\kappa,\lambda}\boldsymbol\delta,\qquad
    \mathbf K^{\kappa,\lambda}=(\mathbf I-\lambda\mathbf D_\kappa\mathbf S)^{-1}.
\end{aligned}
\label{eq:cct_matrix}
\end{equation}
Because $\mathbf S$ is nilpotent on a finite response, the inverse is a finite causal series. Its entries are
\begin{equation}
K^{\kappa,\lambda}_{t,u}
=\ind[u\ge t]\lambda^{u-t}\prod_{j=t}^{u-1}\kappa_j.
\label{eq:cct_kernel}
\end{equation}
Thus the exact local sensitivity is
\begin{equation}
    \frac{\partial A_t}{\partial \delta_u}
    =K^{\kappa,\lambda}_{t,u}.
    \label{eq:sensitivity}
\end{equation}
This equation is the core conceptual object. Standard GAE uses one stationary kernel for all rollouts. \framework uses a policy-computation-conditioned causal kernel whose paths differ across positions and trajectories.

\subsection{Latent continuation interpretation}
An estimator-side latent variable $c_t\in\{0,1\}$ can be used for intuition. If
\begin{equation}
    \Pr(c_t=1\mid\mathcal G_{\theta^-}(h_t))=\kappa_t,
\end{equation}
then marginalizing $c_t$ yields the one-step rule
\begin{equation}
(\mathcal T_\kappa V)(h_t)
=\E[r_t+\kappa_tV(h_{t+1})\mid h_t].
\label{eq:tkappa}
\end{equation}
This does not claim that the environment terminates with probability $1-\kappa_t$. It says that the estimator retains an expected fraction $\kappa_t$ of downstream value across the current computational transition.

Transition-dependent discounting is well established as a way to specify task horizon or temporal abstraction \citep{white2017taskspec,harutyunyan2019optiondiscount}; state-dependent discounting can also be learned with additional consistency controls \citep{wang2026adagamma}. \framework differs in role and coupling: the external task remains fixed, the gate is read from behavior-policy computation, and the same coordinate is carried through the trace and value representation. We therefore use ``retention gate'' rather than treating $\kappa_t$ as a new normative task discount.

\subsection{Outer update and scope}
The actor uses the unchanged clipped surrogate
\begin{equation}
\begin{aligned}
\mathcal L_{\mathrm{clip}}(\theta;\theta^-)
=\E_t\!\big[\min\big(&\rho_t(\theta)\widehat A_t,\\[-1mm]
&\clip(\rho_t(\theta),1-\epsilon,1+\epsilon)\widehat A_t\big)\big].
\end{aligned}
\label{eq:ppo_outer}
\end{equation}
where $\rho_t=\pi_\theta(y_t\mid h_t)/\pi_{\theta^-}(y_t\mid h_t)$ and $\widehat A_t$ is the normalized \compgae estimate. Reward computation, rollout sampling, and KL regularization are unchanged.

This scope avoids two overclaims. First, \framework is not an attention-weighted policy loss: $\kappa_t$ acts inside the bootstrap and trace before the policy objective is formed. Second, we do not claim an unbiased gradient theorem for a current-policy-dependent discounted-return objective. The implemented object is a detached, rollout-conditioned surrogate. Gate staleness across multiple update epochs is a real approximation analyzed as a limitation.

\subsection{The general design space}
The scalar gate studied here is the smallest member of a larger family. A computation-conditioned transport design specifies:
\begin{enumerate}
    \item an internal computation object $\mathcal G$ and summary $\Phi$;
    \item a bounded transport map $g$ or a graph-valued kernel;
    \item a local evidence vector, such as TD residuals, process rewards, tree values, or counterfactual scores;
    \item a transport-aligned baseline or critic; and
    \item an outer update geometry, potentially including computation-conditioned clipping or KL.
\end{enumerate}
Attention concentration is one deployable instantiation, not a claim of optimality. Future variants can use full attention-flow graphs, retrieval provenance, tool-observation edges, MoE expert assignments, recurrent memory gates, or modality-specific routing.

%% file: sections/04_method.tex
\section{\method: An Attention-Routed Instantiation}
\label{sec:method}

\input{figures/comppo_pipeline}

\subsection{Attention concentration as routing sharpness}
For each valid response token $t$, let $\mathcal I_t$ be the set of valid historical positions exposed to the selected causal-attention layer and let $n_t=|\mathcal I_t|$. We aggregate final-layer GQA attention across query heads and renormalize over $\mathcal I_t$ to obtain $a_{ti}\ge 0$ with $\sum_{i\in\mathcal I_t}a_{ti}=1$. The routing-sharpness statistic is the Herfindahl concentration
\begin{equation}
    H_t=\sum_{i\in\mathcal I_t}a_{ti}^2,
    \qquad H_t\in[1/n_t,1].
    \label{eq:herfindahl}
\end{equation}
Uniform attention gives $H_t=1/n_t$ and a point mass gives $H_t=1$. Because the raw lower bound changes with available history, we normalize it as
\begin{equation}
    c_t=\frac{\log(n_tH_t)}{\log n_t}\in[0,1],
    \label{eq:cnorm}
\end{equation}
for $n_t>1$, with $c_t=1/2$ for the first valid position. This transformation maps the two position-dependent extremes exactly to zero and one while preserving ordering. It is not a learned position schedule.

The gate is
\begin{equation}
\kappa_t=\kappamin+(\kappamax-\kappamin)
\sigma\!\left(\tau(c_t-1/2)\right),
\label{eq:kappa_attention}
\end{equation}
with configured envelope parameters $(\kappamin,\kappamax,\tau)=(0.1,0.9,4)$ in all main experiments. Because the sigmoid has finite temperature, the attained range is the strict subinterval $[g(0),g(1)]$. Larger $c_t$ means that the current policy computation is dominated by fewer historical inputs and therefore retains more downstream value across this transition.

The semantic claim is intentionally limited. $H_t$ does not identify which token causally earned the reward. It summarizes the \emph{shape} of upstream support for the current decision. The question answered by $\kappa_t$ is ``how much downstream value should cross this transition under the chosen transport model?'', not ``which past token is a causal explanation?'' The shuffle and position-only controls in \cref{sec:signal-controls} test whether the realized signal carries information beyond a generic position-dependent gate.

\subsection{Computation-aligned critic}
\label{sec:critic}
The value estimator must predict targets generated by \cref{eq:comp_td} while remaining inexpensive enough for 16K-token rollouts. Rather than train a second Transformer, we reuse actor activations and add a small critic with fewer than $0.5\%$ new parameters.

\paragraph{Cross-layer actor feature fusion.}
For a selected set of $L'$ upper layers,
\begin{equation}
    \bar h_t=\sum_{\ell=1}^{L'}\omega_\ell h_t^{(\ell)},
    \qquad \omega=\softmax(\eta),
    \label{eq:fusion}
\end{equation}
so the critic can combine representations at different abstraction levels.

\paragraph{Value-refined history routing.}
Native attention was trained for next-token computation, not scalar value prediction. We therefore modulate each edge with a learned bilinear relevance gate,
\begin{align}
\beta_{ti}&=\frac{\langle W_{\beta,q}\bar h_t,W_{\beta,k}\bar h_i\rangle}{\sqrt d},\\
\widetilde a_{ti}&=a_{ti}\,\sigma(\beta_{ti}).
\label{eq:valuegate}
\end{align}
A second nonnegative token-relevance score
\begin{equation}
    m_{ti}=\operatorname{ReLU}\!\left(
    \frac{\langle W_{r,q}\bar h_t,W_{r,k}\bar h_i\rangle}{\sqrt d}
    \right)
\label{eq:relevance}
\end{equation}
filters superficial routing correlations. The earlier version called this module ``TokenMI''; because no explicit mutual-information bound is optimized, the present paper uses the operationally accurate name \emph{token-relevance gate}. The final causal pooling score and routed-history representation are
\begin{align}
    s_{ti}&=\widetilde a_{ti}m_{ti}\mathbf 1[i\in\mathcal I_t],\\
    h_t^G&=\sum_{i\in\mathcal I_t}
    \frac{s_{ti}}{\sum_{j\in\mathcal I_t}s_{tj}+\varepsilon}\bar h_i.
\label{eq:globalpool}
\end{align}

\paragraph{Gate-aligned local/global value.}
Two lightweight heads produce a local value $V^L(\bar h_t)$ and a routed-history value $V^G(h_t^G)$. We combine them using the same transition regime that generated the bootstrap target:
\begin{equation}
V_\psi(s_t^{\mathrm C})=
\clip\!\left(
\kappa_tV^G(h_t^G)+(1-\kappa_t)V^L(\bar h_t),-1,1
\right).
\label{eq:value_mix}
\end{equation}
This is a co-design hypothesis, not a universal theorem about all critics. Under concentrated routing, the value estimate relies more on explicitly pooled history; under diffuse routing, it relies more on the local actor state. The factorial evidence in \cref{sec:factorial} tests whether this coupling matters empirically while holding the stability stack fixed.

\subsection{Actor and critic objectives}
The actor uses \cref{eq:ppo_outer} with normalized \compgae advantages. The critic regresses to the detached $\lambda$-return
\begin{equation}
    \widehat G_t^{\kappa,\lambda}
    = A_t^{\kappa,\lambda}+V_\psi(s_t^{\mathrm C})m_t
\end{equation}
with clipped value loss. The stored attention-derived gate is detached and cannot be changed by the critic loss within the current rollout batch. The critic consumes actor features from the same forward pass; all reported variants use the same feature-sharing convention so that the controlled contrasts isolate only the gate and critic architecture. The exact backbone-gradient routing is an implementation fact that must be copied from the released code and is flagged in the accompanying author-verification checklist.

\subsection{Stability and systems implementation}
History-conditioned bootstrapping introduces the familiar actor--critic risks of early value error and policy--critic co-divergence. The reported system uses three controls:
\begin{enumerate}
    \item a fixed value-domain clamp $[-1,1]$, matched to the bounded reward scale;
    \item zero-output initialization and a brief critic warm-up, so early advantages are not dominated by arbitrary bootstrap predictions; and
    \item a closed-loop phase controller that adjusts actor learning rate, clip range, value-loss clip, and critic multiplier only when jointly monitored stability statistics cross predefined thresholds.
\end{enumerate}
The core $2\!\times\!2$ experiment holds all three controls identical across cells, so they cannot explain the gate-by-critic interaction.

The implementation avoids materializing the full $T\times T$ attention matrix: only the final layer exposes chunked attention; GQA heads are aggregated by group; buffers are restricted to response positions; online-softmax recomputation avoids dense HBM writes; and top-$K$ causal pooling reduces critic work to $O(TK)$ with $K=64$. These choices avoid a separate same-scale value Transformer and its optimizer state; the additional work is confined to attention extraction and the lightweight critic. Exact measured wall-clock, peak-memory, and throughput records are unavailable in the retained experiment logs. We therefore restrict the present efficiency claim to the algorithmic structure---actor-feature reuse, no second same-scale value Transformer, and $O(TK)$ routed pooling---rather than claiming an unmeasured runtime advantage.

%% file: figures/comppo_pipeline.tex
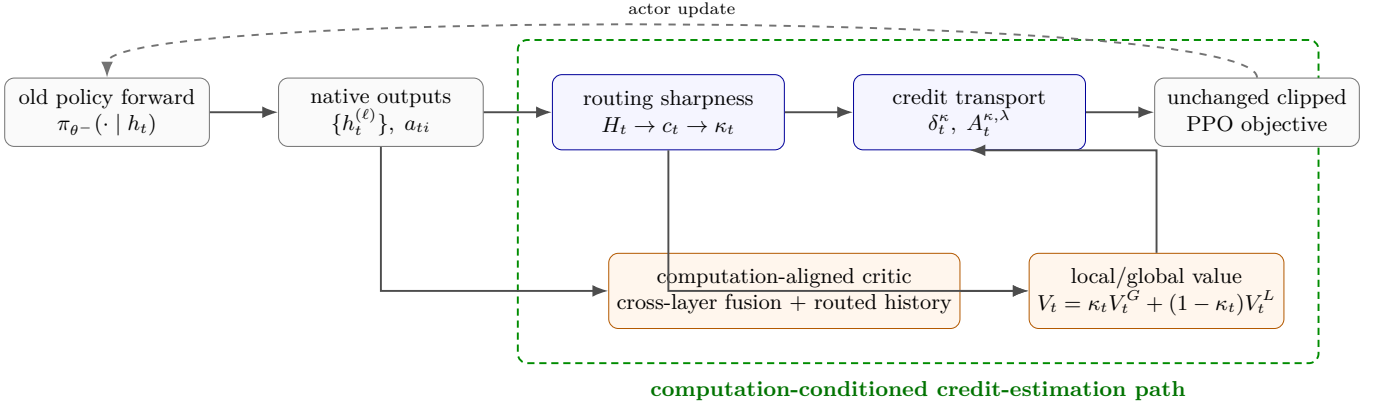
\begin{figure*}[t]
\centering
\resizebox{\textwidth}{!}{%
\begin{tikzpicture}[font=\small,>=Latex,
  block/.style={rounded corners,draw=black!55,fill=black!2,minimum height=10mm,minimum width=30mm,align=center},
  core/.style={rounded corners,draw=blue!60!black,fill=blue!4,minimum height=11mm,minimum width=34mm,align=center},
  critic/.style={rounded corners,draw=orange!70!black,fill=orange!7,minimum height=11mm,minimum width=34mm,align=center},
  arr/.style={->,thick,draw=black!70},
  dashedarr/.style={->,thick,dashed,draw=black!55}]

\node[block] (actor) {old policy forward\\$\pi_{\theta^-}(\cdot\mid h_t)$};
\node[block,right=10mm of actor] (native) {native outputs\\$\{h_t^{(\ell)}\},\;a_{ti}$};
\node[core,right=10mm of native] (conc) {routing sharpness\\$H_t\rightarrow c_t\rightarrow\kappa_t$};
\node[core,right=10mm of conc] (td) {credit transport\\$\delta_t^\kappa,\;A_t^{\kappa,\lambda}$};
\node[block,right=10mm of td] (ppo) {unchanged clipped\\PPO objective};

\node[critic,below=15mm of conc,xshift=17mm] (pool) {computation-aligned critic\\cross-layer fusion + routed history};
\node[critic,right=10mm of pool] (value) {local/global value\\$V_t=\kappa_tV_t^G+(1-\kappa_t)V_t^L$};

\draw[arr] (actor) -- (native);
\draw[arr] (native) -- (conc);
\draw[arr] (conc) -- (td);
\draw[arr] (td) -- (ppo);
\draw[arr] (native.south) |- (pool.west);
\draw[arr] (conc.south) |- (value.west);
\draw[arr] (pool) -- (value);
\draw[arr] (value.north) |- (td.south);
\draw[dashedarr] (ppo.north) .. controls +(0,10mm) and +(0,10mm) .. node[above,font=\scriptsize] {actor update} (actor.north);

\begin{scope}[on background layer]
\node[draw=green!55!black,densely dashed,line width=.8pt,rounded corners,
      fit=(conc)(td)(pool)(value),inner sep=5mm] (creditpath) {};
\end{scope}
\node[font=\bfseries\small,text=green!45!black,anchor=north]
  at ([yshift=-1.2mm]creditpath.south) {computation-conditioned credit-estimation path};
\end{tikzpicture}%
}
\caption{\method changes the credit-estimation path while preserving the task reward and outer clipped policy update. A native policy-computation statistic yields a detached gate $\kappa_t$. The same gate controls one-step bootstrap, eligibility-trace continuation, and the local/global geometry of a critic that reuses actor features.}
\label{fig:pipeline}
\end{figure*}

%% file: sections/05_theory.tex
\section{Properties, Identifiability, and Claim Boundary}
\label{sec:theory}

The results below condition on a fixed rollout batch. The gates were computed by $\pi_{\theta^-}$ and are detached, which is the regime implemented by the algorithm.

\begin{proposition}[Strict reduction]
If $\kappa_t=\gamma$ for every valid transition and the standard critic state is restored, then the computation-conditioned bootstrap, \compgae, and clipped policy update reduce exactly to the standard fixed-$\gamma$ GAE/PPO estimator path.
\end{proposition}

The fixed-gate + aligned-critic ablation is intentionally not vanilla PPO: it preserves the proposed value representation while removing trajectory-specific transport. This is necessary to isolate the gate rather than replacing the entire system at once.

\begin{proposition}[Causal transport kernel]
For a finite response, \cref{eq:comp_gae} has the unique solution
\begin{equation}
A_t^{\kappa,\lambda}
=\sum_{u=t}^{T}
\lambda^{u-t}\left(\prod_{j=t}^{u-1}\kappa_j\right)\delta_u^\kappa.
\label{eq:unrolled}
\end{equation}
Equivalently, $\mathbf A=\mathbf K^{\kappa,\lambda}\boldsymbol\delta$ with entries in \cref{eq:cct_kernel}. If $\kappa_t=\gamma$, the kernel is stationary Toeplitz and equals \cref{eq:toeplitz}; otherwise it is generally path-dependent and nonstationary.
\end{proposition}

The proposition formalizes ``let credit follow computation.'' It does not imply nonlocal token attribution: the method still transports evidence through a scalar sequential kernel. What changes is the local retention assigned to each transition, and therefore the sensitivity of every earlier advantage to every later residual.

\begin{proposition}[Well-posed detached backup]
For a fixed policy and a gate satisfying $0\le\kappa_t\le\bar\kappa<1$, the operator in \cref{eq:tkappa} is a sup-norm contraction with modulus at most $\bar\kappa$:
\begin{equation}
\|\mathcal T_\kappa V-\mathcal T_\kappa W\|_\infty
\le\bar\kappa\|V-W\|_\infty.
\end{equation}
It therefore has a unique bounded fixed point.
\end{proposition}

This is a fixed-policy evaluation statement. It does not establish global convergence of PPO while the behavior policy, critic, and future gates change.

\begin{proposition}[Bounded trace]
If $|\delta_t^\kappa|\le D$ and $\lambda\bar\kappa<1$, then
\begin{equation}
|A_t^{\kappa,\lambda}|
\le \frac{D}{1-\lambda\bar\kappa}.
\end{equation}
\end{proposition}

The trace is therefore controlled by the largest realized gate, not by sequence length directly. TD bootstrapping still provides local residuals: even with $r_t=0$, $\delta_t^\kappa$ can be nonzero whenever the critic detects a value change. We do not claim that terminal reward is transmitted without decay through an arbitrarily long chain.

\paragraph{Transport identifiability.}
A performance gain from dynamic gates could arise from at least four explanations: a better average horizon, a useful position schedule, trajectory-specific alignment, or critic capacity. The experimental design explicitly separates them. The fixed gate is set to the empirical mean $0.61$; a position-only control freezes the mean gate in 20 equal-width position bins; global shuffling preserves the gate values but destroys token correspondence; and the $2\!\times\!2$ factorial crosses transport with critic architecture. These controls do not establish causal token attribution, but they make the trajectory-specific transport claim falsifiable.

\paragraph{Old-policy dependence and staleness.}
The gate is computed from $\theta^-$ and reused during the actor epochs. As $\theta$ moves, its current internal computation may differ from the stored coordinate. This resembles other rollout-derived quantities in PPO but introduces a new source of estimator staleness. The present implementation uses two actor epochs and clipping; a future analysis should bound the error in terms of policy-ratio drift, gate drift, and critic error, or recompute the gate per epoch.

\paragraph{Configured envelope versus realized range.}
The attention instantiation uses the smooth map in \cref{eq:kappa_attention}. With a finite temperature, the configured parameters $(\kappamin,\kappamax)$ are asymptotic envelope parameters; the actual mathematical range is $[g(0),g(1)]\subset(\kappamin,\kappamax)$. All propositions use the actual supremum $\bar\kappa=\sup_c g(c)<1$. Descriptive gate statistics reported later focus on mean scale, dispersion, position, and outcome stratification.

\paragraph{What is not proved.}
We do not prove that attention is a causal explanation, that the chosen concentration statistic is optimal, that every standard critic fails, or that \method dominates tuned PPO in every regime. The theoretical contribution is the transport formulation and its fixed-batch properties; the algorithmic claims are supported by controlled experiments in the stated domain.

%% file: sections/06_experiments.tex
\section{Experimental Design}
\label{sec:experiments}

The experiments are organized around five falsifiable questions rather than a flat list of baselines.
\begin{enumerate}
    \item \textbf{Performance:} does computation-conditioned credit improve over a credible group-relative baseline under a matched search budget?
    \item \textbf{Mechanism:} are the dynamic gate and computation-aligned critic jointly necessary, or is the result explained by either component alone?
    \item \textbf{Signal specificity:} does trajectory-specific gate alignment matter beyond position or marginal regularization?
    \item \textbf{Robustness:} does the method occupy a wider stable region than PPO in a matched stress grid?
    \item \textbf{Transfer and baseline strength:} do the gains survive frozen evaluation, a second backbone, and a larger GRPO group?
\end{enumerate}

\subsection{Task, models, and evaluation}
Training uses hard mathematical-reasoning prompts from DAPO-Math-17K with a maximum response length of 16K. The primary backbone is Qwen3-4B \citep{yang2025qwen3}; Llama-3.1-8B-Instruct \citep{meta2024llama3} is used as a transfer control. A deterministic answer extractor supplies terminal correctness reward. Malformed reasoning tags receive a negative terminal reward, and the length penalty is applied only to correct, format-valid responses.

The in-distribution (ID) development/validation set contains 500 held-out problems and is distinct from frozen MATH500. We use the following endpoint conventions throughout:
\begin{itemize}
    \item \textbf{Best}: maximum ID validation accuracy within the predeclared training budget;
    \item \textbf{Final}: the last checkpoint (step 200 for the performance and factorial experiments; step 150 for the PPO stress test);
    \item \textbf{Last-20}: mean over the final 20 checkpoints;
    \item \textbf{AULC}: validation accuracy integrated over training steps and divided by the observed step range.
\end{itemize}
Final is the primary endpoint. Frozen evaluation always uses the final checkpoint and is never consulted for hyperparameter selection.

Frozen benchmarks are AIME25, AIME2024, MATH500, GSM8K, LiveBench, GPQA, and an Olympiad-style set. We report greedy pass@1 and majority@8. Macro-averages are unweighted across benchmarks and should be interpreted alongside per-benchmark values, especially because AIME sets contain only 30 items.

\subsection{Common training configuration}
All primary Qwen runs use batch size 4, four responses per prompt, PPO/GRPO minibatch size 4, 16K maximum response length, BF16, AdamW with $(\beta_1,\beta_2)=(0.9,0.999)$, zero weight decay, gradient clipping at 1, two update epochs, and sampling temperature/top-$p$ of $1.0/0.7$. The actor learning-rate candidates are $\{5\!\times\!10^{-7},10^{-6},2\!\times\!10^{-6}\}$ and KL candidates are $\{0,10^{-3},2\!\times\!10^{-3}\}$ for both GRPO and \method. The selected GRPO configuration is $(10^{-6},10^{-3})$; the selected \method configuration is $(10^{-6},2\!\times\!10^{-3})$. The \method critic uses learning rate $10^{-5}$, $\lambda=0.95$, configured gate-envelope parameters $(0.1,0.9)$ with $\tau=4$, and value clamp $[-1,1]$.

The same $3\!\times\!3$ grid is screened once on ID validation for each method. The selected configuration is then rerun with five independent seeds $\{17,42,123,256,2026\}$. The submitted seed is 42. At every learning rate, the selected-KL \method endpoint exceeds the selected-KL GRPO endpoint; full grids are reported in \cref{app:fullresults}.

\subsection{Core coefficient-by-critic factorial}
\label{sec:factorial-design}
We cross two transport rules with two critic classes:
\begin{center}
\small
\begin{tabularx}{\columnwidth}{l l >{\raggedright\arraybackslash}X}
\toprule
Transport & Critic & Interpretation\\
\midrule
fixed $\kappa=0.61$ & standard & mean-matched fixed-coordinate control\\
$\kappa_t$ / \compgae & standard & dynamic transport without aligned critic\\
fixed $\kappa=0.61$ & aligned & aligned critic without trajectory gate\\
$\kappa_t$ / \compgae & aligned & full \method\\
\bottomrule
\end{tabularx}
\end{center}
The constant 0.61 equals the empirical mean gate of full \method. In the fixed/aligned cell, it replaces $\kappa_t$ both in the GAE recursion and in the local/global value mixture, removing trajectory dependence while preserving all other architecture and stability components. All four cells share the same Qwen checkpoint, hard-DAPO prompt stream, 500-problem ID set, data seeds, response budget, reward, optimizer, KL/clip configuration, stability system, and 200-step budget.

We report policy endpoints and critic diagnostics on a shared held-out token sample: explained variance (EV), Spearman correlation with TD targets, value MAE, and clamp saturation. The seed-matched factorial interaction is
\begin{equation}
\mathcal I=\text{Full}-(\text{Gate+Std})-(\text{Fixed+Aligned})+(\text{Fixed+Std}).
\label{eq:interaction}
\end{equation}
A positive $\mathcal I$ indicates complementarity beyond additive main effects.

\subsection{Gate controls}
\label{sec:signal-controls-design}
We use two negative controls.
\paragraph{Global shuffle.} Attention and gates are computed normally, then realized $\kappa_t$ values are randomly permuted across valid response positions before both \compgae and critic mixing. This preserves the global marginal distribution but destroys token-wise alignment.
\paragraph{Position-only schedule.} For each seed, full-training rollout gates are averaged within 20 equal-width bins of normalized response position $t/T$. The schedule is frozen and used in both \compgae and critic mixing; no validation or frozen benchmark data are used to construct it. This preserves the mean positional profile but removes trajectory-specific variation.

We also report the realized gate distribution from the final checkpoints, including mean, median, interquartile range, response-position thirds, and correct/incorrect trajectories.

\subsection{Matched PPO stress grid}
The PPO comparison is a robustness stress test, not a claim of universally tuned superiority. PPO and \method are evaluated for 150 steps over actor learning rates $\{2\!\times\!10^{-7},5\!\times\!10^{-7},10^{-6}\}$ and KL coefficients $\{0,10^{-3}\}$, with two seeds per cell. No annealing, extra entropy/KL constraint, critic warm-up, or adaptive controller is used. Both methods use the same fixed value-domain bound $[-1,1]$; separate no-clamp checks show that it acts only during a comparable early critic transient.

A run is declared stable when final accuracy is at least the 42.4\% base accuracy, peak-to-final drop is below 15 points, and there is no sequence of five consecutive post-peak checkpoints below base. We report stable runs, mean best/final, and peak-to-final drop. The permitted inference is a wider stable region in this tested grid; it is not that PPO cannot be stabilized by method-specific tuning.

\subsection{Stronger GRPO group control}
The submitted GRPO uses four prompts and four responses, i.e., 16 trajectories per update. We additionally evaluate a requested $8\!\times\!8$ single-run GRPO control with 64 trajectories per update and all other settings matched. This result is a point estimate without training-seed confidence intervals; it tests baseline sensitivity rather than replacing the five-seed matched-budget comparison.

%% file: sections/07_results.tex
\section{Results}
\label{sec:results}

\input{figures/results_main}

\subsection{Five-seed performance against GRPO}
\label{sec:main-performance}
The selected GRPO configuration obtains $56.2$\% best accuracy (95\% CI $[55.6,56.9]$) and $53.8$\% final accuracy $[52.9,54.7]$. \method obtains $61.7$\% best $[61.3,62.1]$ and $61.4$\% final $[60.8,62.0]$. Every \method seed exceeds every GRPO seed at both endpoints. The method's best-to-final drop is only $0.3$ point, compared with $2.4$ points for GRPO.

The difference is not an isolated selected cell. At actor learning rates $5\!\times\!10^{-7}$, $10^{-6}$, and $2\!\times\!10^{-6}$, each method's preferred-KL endpoint yields \method--GRPO gains of $+6.8/+7.2$, $+5.4/+7.6$, and $+6.8/+8.0$ best/final points, respectively. The complete screen appears in \cref{tab:tuninggrid}. This pattern argues against a single lucky learning-rate/KL choice.

\subsection{The gate and critic form a coupled mechanism}
\label{sec:factorial}
\Cref{fig:factorial_bars,tab:factorial} summarize the controlled $2\!\times\!2$. A fixed mean-matched gate with a standard critic reaches $52.6$\% final. Replacing only the trace by \compgae raises the result to $55.2$\%, but the tested standard critic still has negative EV ($-0.08$), weak TD-target Spearman correlation ($0.16$), and never reaches the submitted GRPO peak of $56.4$\%. Replacing only the critic raises final accuracy to $56.4$\%, with EV $0.45$, but remains five points below full \method. The complete method reaches $61.4$\%, EV $0.52$, and TD-target Spearman $0.57$.

\begin{table*}[t]
\centering
\small
\setlength{\tabcolsep}{4.5pt}
\resizebox{\textwidth}{!}{%
\begin{tabular}{lcccccccccc}
\toprule
Variant & Best & Final & Last-20 & AULC & Peak--final & EV & TD $\rho_s$ & MAE & Clamp sat. & Step to 56.4\\
\midrule
Fixed gate + standard critic & 53.4 & 52.6 & 52.0 & 47.4 & 0.8 & $-0.15$ & 0.11 & 0.41 & 0.24 & never\\
\compgae + standard critic & 56.0 & 55.2 & 54.7 & 50.2 & 0.8 & $-0.08$ & 0.16 & 0.37 & 0.19 & never\\
Fixed gate + aligned critic & 57.2 & 56.4 & 55.9 & 53.1 & 0.8 & 0.45 & 0.52 & 0.20 & 0.05 & 112\\
Full \method & \textbf{61.7} & \textbf{61.4} & \textbf{60.8} & \textbf{58.9} & \textbf{0.3} & \textbf{0.52} & \textbf{0.57} & \textbf{0.18} & \textbf{0.03} & \textbf{33}\\
\bottomrule
\end{tabular}%
}
\caption{Five-seed gate-by-critic factorial. Best and final means have 95\% Student-$t$ intervals: Fixed+Std $53.4[52.9,53.9]/52.6[52.0,53.2]$; \compgae+Std $56.0[55.6,56.5]/55.2[54.6,55.8]$; Fixed+Aligned $57.2[56.6,57.8]/56.4[55.8,57.0]$; Full $61.7[61.3,62.1]/61.4[60.8,62.0]$.}
\label{tab:factorial}
\end{table*}

Seed-matched differences sharpen the attribution. Full \method exceeds \compgae+standard critic by $+5.7$ best and $+6.2$ final points, and exceeds fixed-gate+aligned critic by $+4.5$ and $+5.0$. The interaction in \cref{eq:interaction} is $+1.9[1.3,2.4]$ for best and $+2.4[1.9,2.9]$ for final. Thus the components are super-additive under the controlled configuration: the aligned critic is not a free-standing replacement for the gate, and the gate cannot realize its full benefit through the tested standard value head.

The critic diagnostics support the proposed mechanism rather than merely restating endpoint accuracy. Standard-critic cells have negative EV, large MAE, and high clamp saturation. Aligned-critic cells explain target variation, rank TD targets more faithfully, and saturate less. We therefore interpret the standard critic result as an empirical target/representation mismatch in this setting, not as a theorem that every conventional critic must fail.

\subsection{Trajectory-specific alignment matters}
\label{sec:signal-controls}
\input{figures/results_controls}
Global shuffling reduces best/final accuracy from $61.7/61.4$ to $59.4/58.7$, with a $2.7$-point final penalty. The position-only schedule recovers part of the effect, reaching $60.2/59.7$. It exceeds the matched fixed gate by $+3.0/+3.3$ best/final points and exceeds global shuffle by $+0.8/+1.0$, showing that the mean positional profile is informative. Nevertheless, full \method exceeds position-only by a paired $+1.5[1.3,1.7]$ best and $+1.7[1.5,1.8]$ final points. The remaining gain therefore depends on trajectory-specific variation rather than position alone.

The realized gate is not constant. Across five final checkpoints, its mean/median are $0.61/0.63$ and its interquartile range is $0.18$. Mean gates in the early, middle, and late response thirds are $0.52/0.63/0.68$, and correct trajectories have higher mean gate than incorrect trajectories ($0.64$ versus $0.57$). These descriptive differences do not establish causality; they show that the retained coordinate varies meaningfully across positions, trajectories, and outcomes.

\subsection{A wider stable region than PPO in the tested grid}
\label{sec:ppo-results}
Across 12 runs, PPO is stable in $3/12$, with mean best/final/drop $47.7/34.4/13.3$. \method is stable in $10/12$, with $52.9/45.6/7.3$. The largest difference occurs at aggressive learning rates: at $10^{-6}$ with KL $10^{-3}$, PPO obtains $49.8/22.6$ best/final while \method obtains $57.2/44.6$; in the originally submitted cell, the corresponding single-seed endpoints were PPO $47.6/15.0$ and \method $58.4/44.2$.

The grid also bounds the conclusion. At the conservative $2\!\times\!10^{-7}$ learning rate, PPO is stable in one of two seeds for each KL value and can finish above base. At $10^{-6}$ with no KL, one \method cell is also unstable. The evidence therefore supports a wider tested stable region and lower average degradation, not universal dominance over a method-specifically tuned PPO.

\subsection{Frozen transfer across benchmarks and backbones}
\label{sec:frozen-results}
\begin{table}[t]
\centering
\small
\setlength{\tabcolsep}{3pt}
\begin{tabular}{lrrrr}
\toprule
Backbone & Base & GRPO & \method & $\Delta$\\
\midrule
Qwen3-4B, greedy & 58.2 & 57.5 & \textbf{61.7} & +4.3\\
Qwen3-4B, majority@8 & 69.7 & 68.7 & \textbf{71.0} & +2.3\\
Llama-3.1-8B, greedy & 31.1 & 39.0 & \textbf{42.8} & +3.9\\
Llama-3.1-8B, majority@8 & 36.5 & 47.8 & \textbf{52.0} & +4.2\\
\bottomrule
\end{tabular}
\caption{Unweighted frozen-benchmark macro-averages. $\Delta$ is \method minus GRPO. These are final-checkpoint point estimates; complete per-benchmark results are in \cref{tab:qwenfrozen,tab:llamafrozen}.}
\label{tab:frozenmacro}
\end{table}

On Qwen3-4B, \method exceeds GRPO by $4.3$ greedy and $2.3$ majority@8 macro points. AIME25 greedy is the exception: GRPO scores $50.0$ while \method and base score $43.3$; majority@8 ties at $66.7$. The strongest Qwen gains occur on AIME2024 ($+16.7$ greedy), GPQA ($+11.6$), LiveBench ($+3.0$), and MATH500 ($+3.2$). On Llama-3.1-8B-Instruct, all seven greedy differences favor \method, for a $+3.9$ macro improvement.

The Qwen GRPO frozen greedy macro is $0.7$ below base despite a $+11.4$-point final improvement on ID validation. This is a small transfer trade-off, not failed optimization. The Llama control further shows that GRPO is a legitimate learning baseline: it improves frozen greedy macro from $31.1$ to $39.0$.

\subsection{Stronger prompt--group GRPO control}
Increasing GRPO from $4\!\times\!4$ to $8\!\times\!8$ improves ID best/final from $56.4/53.8$ to $58.0/56.2$ and frozen greedy macro from $57.5$ to $60.2$. The corresponding submitted-seed \method reference is $61.8/61.4$ ID and $61.7$ frozen macro. Thus the stronger group closes the frozen gap to $1.5$ points but leaves descriptive gaps of $+3.8$ best and $+5.2$ final ID points. Because this control has one training run, we present it as baseline-sensitivity evidence rather than a significance-tested comparison.

\paragraph{Summary.}
The evidence chain is consistent across levels: matched tuning establishes performance; the factorial isolates coupling; shuffled and position-only controls isolate trajectory-specific alignment; target metrics support the critic interpretation; the stress grid bounds robustness; and frozen/transfer controls show that the effect is not confined to a single ID curve. No individual experiment establishes universal generality, but together they support computation-conditioned credit as a substantive algorithmic variable rather than an attention-themed engineering add-on.

%% file: figures/results_main.tex
\begin{figure*}[t]
\centering
\begin{subfigure}[t]{0.48\textwidth}
\centering
\begin{tikzpicture}
\begin{axis}[
    ybar,bar width=9pt,width=\linewidth,height=5.2cm,
    ymin=48,ymax=64,
    ylabel={ID validation accuracy (\%)},
    symbolic x coords={GRPO,CompPO},xtick=data,
    legend style={at={(0.5,1.03)},anchor=south,legend columns=2,draw=none},
    legend image code/.code={\draw[#1] (0cm,-0.09cm) rectangle (0.25cm,0.09cm);},
    nodes near coords,nodes near coords style={font=\scriptsize},
    enlarge x limits=0.35,
    error bars/y dir=both,error bars/y explicit]
\addplot+[fill=black!25,draw=black!65] coordinates {(GRPO,56.2) +- (0,0.65) (CompPO,61.7) +- (0,0.4)};
\addplot+[fill=blue!45,draw=blue!70!black] coordinates {(GRPO,53.8) +- (0,0.9) (CompPO,61.4) +- (0,0.6)};
\legend{Best,Final}
\end{axis}
\end{tikzpicture}
\caption{Five-seed selected configurations; bars show means and 95\% Student-$t$ intervals.}
\label{fig:main_endpoints}
\end{subfigure}\hfill
\begin{subfigure}[t]{0.50\textwidth}
\centering
\begin{tikzpicture}
\begin{axis}[
    ybar,bar width=13pt,width=\linewidth,height=5.2cm,
    ymin=49,ymax=64,
    ylabel={Final ID accuracy (\%)},
    symbolic x coords={Fixed+Std,Gate+Std,Fixed+Aligned,Full},
    xtick=data,xticklabel style={rotate=20,anchor=east,font=\scriptsize},
    nodes near coords,nodes near coords style={font=\scriptsize},
    enlarge x limits=0.13,
    error bars/y dir=both,error bars/y explicit]
\addplot+[fill=orange!42,draw=orange!80!black] coordinates {
(Fixed+Std,52.6) +- (0,0.6)
(Gate+Std,55.2) +- (0,0.6)
(Fixed+Aligned,56.4) +- (0,0.6)
(Full,61.4) +- (0,0.6)};
\end{axis}
\end{tikzpicture}
\caption{Gate-by-critic factorial. The final interaction is $+2.4$ points $[1.9,2.9]$.}
\label{fig:factorial_bars}
\end{subfigure}
\caption{Main training evidence. \method improves both best and final endpoints over the selected GRPO baseline, and the factorial experiment shows that the trajectory-specific gate and aligned critic are complementary rather than interchangeable.}
\label{fig:main_results}
\end{figure*}
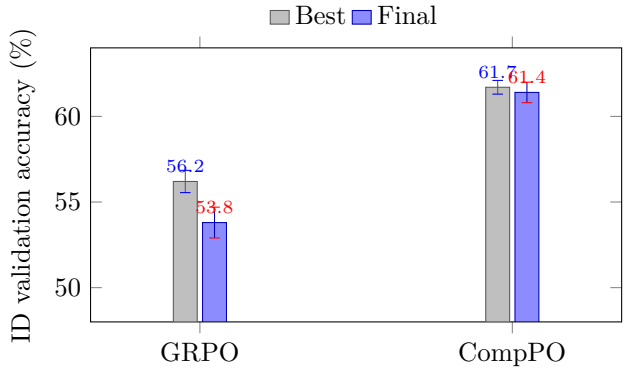
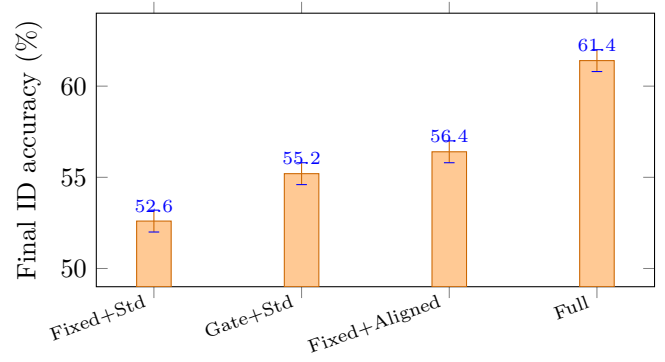

%% file: figures/results_controls.tex
\begin{figure*}[t]
\centering
\begin{subfigure}[t]{0.48\textwidth}
\centering
\begin{tikzpicture}
\begin{axis}[
    ybar,bar width=14pt,width=\linewidth,height=5.0cm,
    ymin=56,ymax=63,
    ylabel={Final ID accuracy (\%)},
    symbolic x coords={Shuffle,Position,Full},xtick=data,
    nodes near coords,nodes near coords style={font=\scriptsize},
    enlarge x limits=0.25,
    error bars/y dir=both,error bars/y explicit]
\addplot+[fill=green!40,draw=green!55!black] coordinates {
(Shuffle,58.7) +- (0,0.7)
(Position,59.7) +- (0,0.5)
(Full,61.4) +- (0,0.6)};
\end{axis}
\end{tikzpicture}
\caption{Gate controls. Full uses trajectory-specific alignment.}
\label{fig:gate_controls}
\end{subfigure}\hfill
\begin{subfigure}[t]{0.48\textwidth}
\centering
\begin{tikzpicture}
\begin{axis}[
    ybar,bar width=18pt,width=\linewidth,height=5.0cm,
    ymin=0,ymax=12,
    ylabel={Stable runs (out of 12)},
    symbolic x coords={PPO,CompPO},xtick=data,
    nodes near coords,nodes near coords style={font=\small},
    enlarge x limits=0.45]
\addplot+[fill=red!35,draw=red!70!black] coordinates {(PPO,3) (CompPO,10)};
\end{axis}
\end{tikzpicture}
\caption{Matched LR--KL stress grid; this is a bounded robustness claim, not tuned PPO ranking.}
\label{fig:robustness}
\end{subfigure}
\caption{Specificity and robustness controls. Destroying token-wise alignment or replacing it with a fixed positional profile reduces performance, while \method occupies a wider stable region in the tested PPO grid.}
\label{fig:controls}
\end{figure*}
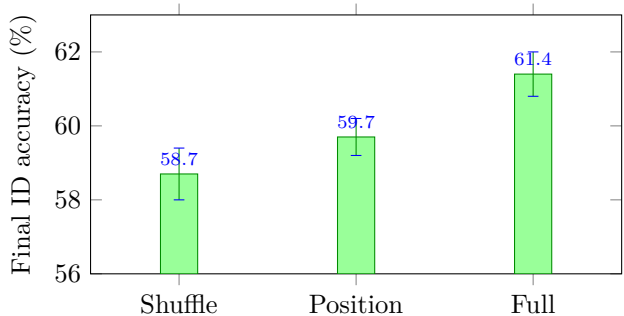
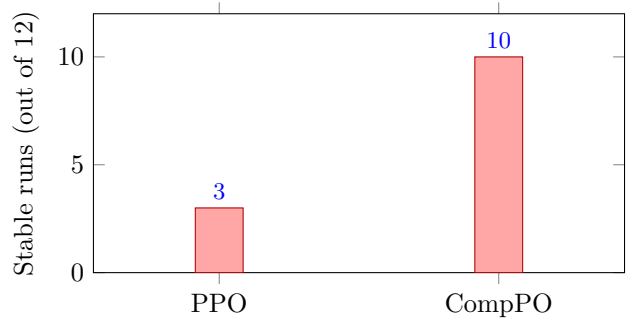

%% file: sections/08_related_work.tex
\section{Related Work}
\label{sec:related}

\paragraph{Fixed and transition-dependent credit in RL.}
TD learning, eligibility traces, GAE, and PPO provide the classical basis for the estimator used here \citep{schulman2015gae,schulman2017ppo}. Generalized RL task formulations allow transition-dependent discounting \citep{white2017taskspec}, and option-level work has studied per-decision discount factors as temporal-abstraction and bias--variance controls \citep{harutyunyan2019optiondiscount}. AdaGamma, developed concurrently, learns state-dependent discounts with a return-consistency objective and analyzes the resulting operator \citep{wang2026adagamma}. \method differs in interpretation and source: the task reward is unchanged, the gate is a detached statistic of the LLM policy's own computation, and the contribution is its joint use in bootstrap, trace, and critic representation.

\paragraph{LLM policy optimization.}
PPO-based RLHF trains actor and value models with clipped policy updates and GAE \citep{ziegler2020finetuning,stiennon2020summarize,ouyang2022instructgpt}. RLOO and GRPO remove the learned value model through leave-one-out or group-relative baselines \citep{ahmadian2024rloo,shao2024deepseekmath}. DAPO, Dr.~GRPO, and GSPO improve large-scale training through asymmetric clipping, dynamic sampling, bias correction, loss normalization, or sequence-level importance ratios \citep{yu2025dapo,liu2025drgrpo,zheng2025gspo}. These methods primarily change baseline statistics, sampling, or update geometry. Computation-conditioned credit changes the Bellman/advantage transport path and can in principle be combined with those advances.

\paragraph{Fine-grained LLM credit.}
Process supervision supplies step-level evidence but requires labels or learned reward models \citep{uesato2022process,lightman2023prm}; PRIME learns implicit process rewards online from outcome labels \citep{cui2025prime}. VinePPO estimates intermediate state values through additional continuations and documents severe weaknesses of conventional value networks on reasoning trajectories \citep{kazemnejad2024vineppo}. VC-PPO attributes long-CoT PPO collapse to value initialization and reward-signal decay and uses value pretraining plus decoupled GAE \citep{yuan2025vcppo}. TEMPO constructs prefix trees and provides critic-free TD corrections at branching points \citep{tran2025tempo}; GRPO-$\lambda$ introduces a critic-free eligibility-trace approximation \citep{parthasarathi2025grpolambda}. OAR reshapes group advantages using token perturbation or gradient-based outcome influence \citep{li2026oar}. High-entropy-token methods update only likely reasoning forks \citep{wang2025entropyminority}. OPPO derives a Bayesian token-level value recursion using an oracle-conditioned likelihood ratio \citep{li2026oppo}. S-trace introduces critic-free selective eligibility traces \citep{mou2026strace}. These works reinforce the importance of nonuniform token credit, but use external rollouts, branch statistics, outcome influence, entropy, oracle scores, or critic-free trace surrogates rather than one policy-internal statistic coupled across bootstrap, trace, and critic. A recent survey organizes this rapidly expanding landscape across reasoning and agentic settings \citep{zhang2026creditsurvey}.

\paragraph{Critics for LLM reasoning.}
The difficulty and cost of LLM value models motivated the shift toward critic-free methods. Recent work reopens this design space from complementary directions. AsyPPO uses lightweight prompt-sharded mini-critics \citep{liu2025asyppo}; Generative Actor--Critic replaces one-shot scalar prediction with a reasoning critic and in-context conditioning \citep{shan2026genac}; VC-PPO improves initialization and target construction \citep{yuan2025vcppo}. \method instead asks whether the critic's \emph{state geometry} matches the policy-internal transport rule. Its factorial result supports the narrower claim that the tested standard head does not track the computation-conditioned target, while an aligned actor-feature critic does.

\paragraph{Attention and internal signals as optimization primitives.}
Attention is a central Transformer mechanism \citep{vaswani2017attention}, but its faithfulness as a causal explanation is debated \citep{jain2019attention,wiegreffe2019attention,serrano2019interpretable}. We use it as an observable routing signal, not as proof of causal attribution. HICRA identifies attention-based planning and anchor tokens and targets their updates \citep{li2025attentionilluminates}; Reinforced Attention Learning directly optimizes internal attention policies for multimodal grounding \citep{li2026ral}. FlowTracer, posted after the initial submission, builds an answer-targeted attention-flow DAG and uses throughput for reward shaping \citep{dong2026flowtracer}. These methods are close in spirit because they treat internal computation as useful for RL. The distinctive object in \method is the \emph{transition-level credit-retention gate}: the same native statistic parameterizes the one-step value backup, eligibility trace, and critic's local/global state representation while leaving reward and outer policy loss unchanged.

\begin{table*}[t]
\centering
\scriptsize
\setlength{\tabcolsep}{3.2pt}
\begin{tabularx}{\textwidth}{>{\raggedright\arraybackslash}p{0.12\textwidth}>{\raggedright\arraybackslash}X>{\raggedright\arraybackslash}X>{\raggedright\arraybackslash}X>{\raggedright\arraybackslash}p{0.17\textwidth}}
\toprule
Family & Evidence or baseline & Transport geometry & Update or value mechanism & Relation to this work\\
\midrule
PPO/GAE & learned scalar value & stationary temporal trace & token-ratio clipping; critic & fixed-kernel ancestor\\
GRPO/RLOO & group or leave-one-out outcome & response-level broadcast & critic-free token update & outcome-coordinate control\\
\makecell[l]{DAPO /\\ Dr.~GRPO /\\ GSPO} & improved sampling and normalization & mainly response/group credit & clipping, loss normalization, or sequence ratios & orthogonal optimizer advances\\
PRIME/PRM & process or implicit step evidence & reward placement plus downstream estimator & process/reward model & complementary evidence axis\\
\makecell[l]{VinePPO /\\ TEMPO} & continuation or prefix-tree values & rollout- or tree-based TD & Monte Carlo or nonparametric value & alternative fine-grained transport\\
\makecell[l]{OAR /\\ S-trace /\\ GRPO-$\lambda$} & influence, entropy, or trace proxy & reshaping or selective traces & critic-free update & neighboring token-credit methods\\
\method & native policy computation & path-dependent Bellman/GAE kernel & actor-feature \tac; clipped PPO & coupled architecture-aware transport\\
\bottomrule
\end{tabularx}
\caption{Positioning by the evidence--transport--update factorization. Categories identify the primary intervention, not an exclusive taxonomy; many systems combine multiple axes.}
\label{tab:related_axes}
\end{table*}

\paragraph{Novelty boundary.}
We therefore do not claim the first use of attention or the first token-level credit method. The contribution is the coupled construction and the broader abstraction it instantiates: policy-internal computation can be an explicit coordinate of Bellman-style credit transport. Concurrent work makes this boundary more, not less, important by showing that internal-signal credit is becoming a distinct research axis.

%% file: sections/09_discussion.tex
\section{Discussion: Toward Architecture-Aware RL}
\label{sec:discussion}

\subsection{The scientific claim is larger than an attention heuristic}
The most conservative reading of \method is a new credit estimator that improves long-chain mathematical RL on two backbones. The more consequential reading is that it exposes a design variable that conventional LLM RL leaves implicit. A Transformer is not merely a large black-box policy with token actions; it computes a trajectory-specific internal routing structure at every decision. Once that structure is made available to the estimator, fixed temporal distance and response-level outcome are no longer the only viable coordinates for transporting credit.

This perspective changes the algorithm-design question from
\begin{quote}
``Which black-box policy optimizer should be applied to an LLM?''
\end{quote}
to
\begin{quote}
``Which internal computation of this policy should parameterize its value transport, baseline, trust region, and update?''
\end{quote}
The latter is an architecture-aware RL question. Attention concentration is one answer, chosen because it is native, bounded after normalization, inexpensive to extract, and falsifiable through shuffling. It is unlikely to be the final answer.

\subsection{Reinterpreting the apparent failure of actor--critic LLM RL}
The rise of critic-free methods has often been read as evidence that value learning is a poor fit for LLM reasoning. The evidence in this paper suggests a more specific diagnosis. The standard critic in the factorial experiment fails to track the computation-conditioned target: EV is negative, TD-target correlation is low, and policy gains stall. The aligned critic tracks the target and enables the full algorithm. This does not prove that one architecture is universally necessary, but it supports a field-level hypothesis:
\begin{quote}
Conventional LLM critics may fail partly because their representation and training target are mismatched to the policy's history routing, not because token-level actor--critic learning is intrinsically unsuitable.
\end{quote}
This distinction matters. If the critic-free turn is driven by a contingent representation failure, then improved value models, generative critics, retrieval-aware critics, and computation-aligned heads may recover the sample-efficiency benefits of TD learning without a second same-scale model.

\subsection{An orthogonal primitive rather than another monolithic optimizer}
LLM RL algorithms often combine several axes in one recipe. It is useful to separate them:
\begin{itemize}
    \item \textbf{reward evidence}: outcome verifiers, process rewards, preference models, or judges;
    \item \textbf{sampling and baseline}: groups, leave-one-out statistics, dynamic sampling, or trees;
    \item \textbf{credit transport}: fixed GAE, counterfactual values, tree TD, oracle recursion, or computation-conditioned traces;
    \item \textbf{update geometry}: token/sequence ratios, clipping, KL, entropy, and normalization;
    \item \textbf{systems}: rollout scheduling, kernels, memory, and parallelism.
\end{itemize}
\method intervenes primarily in the third axis and co-designs the critic representation needed by that intervention. This makes it potentially composable with DAPO-style clipping, GSPO-style sequence ratios, process rewards, prefix-tree values, or larger group baselines. The present experiments deliberately keep the outer objective unchanged to isolate the new axis.

\subsection{From interpretability signals to optimization primitives}
Most uses of attention ask whether it explains a prediction. Computation-conditioned credit asks a different question: whether an internal signal can improve the estimator used to train the policy. The burden of proof is therefore empirical alignment, not human interpretability. A signal may be useful for optimization even when it is not a faithful natural-language explanation, provided negative controls show that its trajectory-specific structure matters and its failure modes are bounded.

This shift extends beyond attention. Future policies expose increasingly explicit internal mechanisms:
\begin{itemize}
    \item MoE models choose experts and routing weights;
    \item retrieval-augmented models select documents and spans;
    \item agents read memory, tool observations, and environment states;
    \item multimodal models route between text, image, audio, and video tokens;
    \item state-space and recurrent architectures expose gates and memory updates.
\end{itemize}
Each mechanism can define a candidate computational coordinate. A mature architecture-aware RL stack may use different coordinates for value transport, exploration, KL control, and critic state construction.

\subsection{From a scalar gate to a computational Bellman graph}
The scalar $\kappa_t$ is intentionally minimal: it compresses the geometry of the upstream computation into one transition coefficient. A richer extension would transport credit directly over a graph. Let $G_t$ contain historical tokens or latent states as nodes and normalized policy-computation weights as edges. A graph Bellman operator could propagate value along multiple nonlocal routes, while a graph critic could estimate node- or subgraph-level value. Multi-head traces could maintain separate horizons for specialist heads; hierarchical traces could operate at token, reasoning-step, and tool-call scales. The present result should be interpreted as feasibility evidence for this larger program, not as proof that scalar concentration is optimal.

\subsection{Where the current evidence stops}
The experiments establish the mechanism on mathematical reasoning, two Transformer backbones up to 8B, terminal verifiers, and 16K responses. They do not establish gains in dialogue, tool use, code, retrieval, multimodal trajectories, dense-reward tasks, or larger models. Those domains are motivated because their dependencies are sparse and nonlocal, but they remain untested.

Attention concentration is a structural proxy. Attention sinks, positional bias, induction heads, and output-layer calibration can corrupt it. The learned relevance gate and negative controls reduce but do not eliminate these risks. The gate is also stale across actor epochs, and the current theory does not provide a global policy-improvement bound for policy-generated gates. Finally, the stronger $8\!\times\!8$ GRPO control and frozen benchmark values are point estimates, while exact measured hardware and wall-clock records are unavailable. These limitations motivate targeted follow-up rather than weakening the central observation that policy-internal computation can carry useful credit information.

%% file: sections/10_conclusion.tex
\section{Conclusion}
\label{sec:conclusion}

LLM policies compute with history, but their credit estimators usually do not. This paper identifies that architecture--credit mismatch and introduces computation-conditioned credit: a detached statistic of the policy's own internal computation controls how downstream value crosses a transition. \method instantiates the idea with attention concentration, a path-dependent GAE trace, and a computation-aligned actor-feature critic, while preserving the external reward and clipped PPO update.

The empirical evidence supports the mechanism at several levels. Under matched tuning and five seeds, \method substantially exceeds GRPO; a controlled gate-by-critic factorial reveals a positive interaction and better critic target tracking; shuffle and position-only controls show that trajectory-specific alignment matters; a PPO stress grid shows a wider stable region; and frozen evaluation transfers across Qwen3-4B and Llama-3.1-8B-Instruct. The contribution is therefore not that attention is universally causal or that one optimizer replaces all others. It is that internal computation can become a first-class variable in RL credit transport. This opens a broader agenda in which attention, routing, retrieval, memory, and modality structure shape Bellman operators, eligibility traces, critics, and trust regions for the policies that actually use them.

%% file: sections/A_algorithm.tex
\section{Full Algorithm}
\label{app:algorithm}

\begin{algorithm}[H]
\caption{Computation-Conditioned Policy Optimization (\method)}
\label{alg:comppo}
\begin{algorithmic}[1]
\Require actor $\pi_\theta$, reference policy $\pi_{\rm ref}$, aligned critic $V_\psi$, verifier $R$, gate parameters $(\kappamin,\kappamax,\tau)$, GAE $\lambda$, clip $\epsilon$
\For{rollout iteration $k=1,2,\ldots$}
    \State Set behavior parameters $\theta^-\leftarrow\theta$
    \State Sample prompts and responses $y_{1:T}\sim\pi_{\theta^-}$; compute rewards $r_{1:T}$ and valid-token masks $m_{1:T}$
    \For{each valid response token $t$}
        \State Extract final-layer GQA-group attention $a_{ti}$ over valid history $\mathcal I_t$ and actor hidden states $\{h_t^{(\ell)}\}$
        \State Renormalize $a_{ti}$ over $i\in\mathcal I_t$; set $n_t=|\mathcal I_t|$
        \State $H_t\leftarrow\sum_{i\in\mathcal I_t}a_{ti}^2$
        \State $c_t\leftarrow\log(n_tH_t)/\log n_t$ if $n_t>1$, else $1/2$
        \State $\kappa_t\leftarrow\kappamin+(\kappamax-\kappamin)\sigma(\tau(c_t-1/2))$
        \State Detach and store $\kappa_t$, attention summaries, and required actor features
        \State Fuse actor features $\bar h_t\leftarrow\sum_\ell\omega_\ell h_t^{(\ell)}$
        \State Compute value-refined scores $s_{ti}$ by \cref{eq:valuegate,eq:relevance}
        \State Pool routed history $h_t^G$ by \cref{eq:globalpool}
        \State Evaluate $V_t\leftarrow\clip(\kappa_tV^G(h_t^G)+(1-\kappa_t)V^L(\bar h_t),-1,1)$
    \EndFor
    \State Set $A_{T+1}=0$ and $V_{T+1}=0$
    \For{$t=T,T-1,\ldots,1$}
        \State $\delta_t\leftarrow r_t+\kappa_tV_{t+1}m_{t+1}-V_tm_t$
        \State $A_t\leftarrow\delta_t+\lambda\kappa_tA_{t+1}m_{t+1}$
        \State $G_t\leftarrow A_t+V_tm_t$
    \EndFor
    \State Normalize valid-token advantages $\widehat A_t$
    \For{each optimization epoch}
        \State Update actor with clipped objective in \cref{eq:ppo_outer} plus reference-policy KL
        \State Update critic by clipped regression to detached $G_t$
        \State Apply the predeclared stability phase controller if enabled
    \EndFor
\EndFor
\end{algorithmic}
\end{algorithm}

\paragraph{Gradient convention.}
The behavior-policy attention and $\kappa_t$ are detached for the whole rollout batch. The critic consumes actor features from the same forward computation and cannot alter the detached, stored transport gate for the current batch. The exact backbone-gradient routing must match the released implementation; all controlled variants use the same convention. The outer actor update consumes detached normalized advantages, as in ordinary PPO.

\paragraph{Fixed-gate controls.}
For a fixed-gate ablation, every occurrence of $\kappa_t$ in the TD residual, trace, and value mixture is replaced by the same constant. The main factorial uses $\kappa=0.61$, the empirical mean of the full method, to avoid confounding trajectory dependence with mean scale.

%% file: sections/B_full_results.tex
\clearpage
\section{Complete Experimental Tables}
\label{app:fullresults}

\subsection{Common tuning screen}
\begin{table}[H]
\centering
\small
\begin{tabular}{llrrr}
\toprule
Method & Actor LR & KL & Best & Final\\
\midrule
GRPO & $5\mathrm e{-7}$ & 0 & 52.8 & 51.2\\
GRPO & $5\mathrm e{-7}$ & $1\mathrm e{-3}$ & 54.0 & 52.4\\
GRPO & $5\mathrm e{-7}$ & $2\mathrm e{-3}$ & 53.2 & 51.8\\
GRPO & $1\mathrm e{-6}$ & 0 & 54.6 & 52.8\\
GRPO & $1\mathrm e{-6}$ & $1\mathrm e{-3}$ & \textbf{56.4} & \textbf{53.8}\\
GRPO & $1\mathrm e{-6}$ & $2\mathrm e{-3}$ & 55.0 & 53.2\\
GRPO & $2\mathrm e{-6}$ & 0 & 53.8 & 51.6\\
GRPO & $2\mathrm e{-6}$ & $1\mathrm e{-3}$ & 54.2 & 52.0\\
GRPO & $2\mathrm e{-6}$ & $2\mathrm e{-3}$ & 53.0 & 51.4\\
\midrule
\method & $5\mathrm e{-7}$ & 0 & 59.8 & 58.4\\
\method & $5\mathrm e{-7}$ & $1\mathrm e{-3}$ & 60.4 & 59.2\\
\method & $5\mathrm e{-7}$ & $2\mathrm e{-3}$ & 60.8 & 59.6\\
\method & $1\mathrm e{-6}$ & 0 & 60.6 & 59.0\\
\method & $1\mathrm e{-6}$ & $1\mathrm e{-3}$ & 61.2 & 60.2\\
\method & $1\mathrm e{-6}$ & $2\mathrm e{-3}$ & \textbf{61.8} & \textbf{61.4}\\
\method & $2\mathrm e{-6}$ & 0 & 59.6 & 58.0\\
\method & $2\mathrm e{-6}$ & $1\mathrm e{-3}$ & 60.8 & 59.4\\
\method & $2\mathrm e{-6}$ & $2\mathrm e{-3}$ & 61.0 & 60.0\\
\bottomrule
\end{tabular}
\caption{Single-seed common $3\!\times\!3$ screen used to select the five-seed configurations. Frozen benchmarks were not used.}
\label{tab:tuninggrid}
\end{table}

\subsection{Five-seed confirmation}
\begin{table}[H]
\centering
\small
\begin{tabular}{rrrrr}
\toprule
Seed & GRPO Best & GRPO Final & \method Best & \method Final\\
\midrule
17 & 55.6 & 52.9 & 61.2 & 60.8\\
42 & 56.4 & 53.8 & 61.8 & 61.4\\
123 & 55.8 & 53.4 & 61.5 & 61.2\\
256 & 56.6 & 54.2 & 61.9 & 61.6\\
2026 & 56.8 & 54.7 & 62.1 & 62.0\\
\midrule
Mean & 56.2 & 53.8 & 61.7 & 61.4\\
95\% CI & [55.6,56.9] & [52.9,54.7] & [61.3,62.1] & [60.8,62.0]\\
\bottomrule
\end{tabular}
\caption{Five-seed selected-configuration confirmation.}
\label{tab:mainseeds}
\end{table}

\subsection{Per-seed factorial results}
\begin{table}[H]
\centering
\scriptsize
\setlength{\tabcolsep}{3pt}
\begin{tabular}{r|rr|rr|rr|rr}
\toprule
& \multicolumn{2}{c|}{Fixed+Std} & \multicolumn{2}{c|}{\compgae+Std} & \multicolumn{2}{c|}{Fixed+Aligned} & \multicolumn{2}{c}{Full}\\
Seed & Best & Final & Best & Final & Best & Final & Best & Final\\
\midrule
17 & 52.8 & 51.9 & 55.6 & 54.6 & 56.6 & 55.7 & 61.2 & 60.8\\
42 & 53.8 & 53.1 & 56.6 & 55.8 & 57.8 & 56.9 & 61.8 & 61.4\\
123 & 53.2 & 52.4 & 55.8 & 54.9 & 57.0 & 56.1 & 61.5 & 61.2\\
256 & 53.6 & 52.8 & 56.2 & 55.4 & 57.4 & 56.6 & 61.9 & 61.6\\
2026 & 53.6 & 52.8 & 56.0 & 55.3 & 57.2 & 56.7 & 62.1 & 62.0\\
\bottomrule
\end{tabular}
\caption{Seed-level gate-by-critic factorial endpoints.}
\label{tab:factorialseeds}
\end{table}

\subsection{Gate controls and descriptive statistics}
\begin{table}[H]
\centering
\small
\begin{tabular}{lrrrr}
\toprule
Variant & Best & Final & Last-20 & AULC\\
\midrule
Global shuffle & 59.4 [58.9,59.9] & 58.7 [58.0,59.4] & 58.1 & 57.2\\
Position-only & 60.2 [59.8,60.6] & 59.7 [59.2,60.2] & -- & --\\
Full \method & 61.7 [61.3,62.1] & 61.4 [60.8,62.0] & 60.8 & 58.9\\
\bottomrule
\end{tabular}
\caption{Five-seed gate controls. Brackets are 95\% Student-$t$ intervals.}
\label{tab:gatecontrols}
\end{table}

\begin{table}[H]
\centering
\small
\begin{tabular}{lr@{\qquad}lr}
\toprule
Statistic & Value & Statistic & Value\\
\midrule
Mean / median & 0.61 / 0.63 & IQR & 0.18\\
Early / middle / late & 0.52 / 0.63 / 0.68 & Correct / incorrect & 0.64 / 0.57\\
\bottomrule
\end{tabular}
\caption{Realized gate statistics from five final full-method checkpoints. Archived tail and boundary-proximity summaries are excluded pending reconciliation of the exact implementation map with the printed smooth-map equation.}
\label{tab:gatestats}
\end{table}

\subsection{Matched PPO stress grid}
\begin{table}[H]
\centering
\small
\begin{tabular}{lllrrc}
\toprule
Method & Actor LR & KL & Best & Final & Stable seeds\\
\midrule
PPO & $2\mathrm e{-7}$ & 0 & 48.2 & 44.8 & 1/2\\
PPO & $2\mathrm e{-7}$ & $1\mathrm e{-3}$ & 49.0 & 46.2 & 1/2\\
PPO & $5\mathrm e{-7}$ & 0 & 46.8 & 31.0 & 0/2\\
PPO & $5\mathrm e{-7}$ & $1\mathrm e{-3}$ & 48.4 & 43.6 & 1/2\\
PPO & $1\mathrm e{-6}$ & 0 & 44.2 & 18.4 & 0/2\\
PPO & $1\mathrm e{-6}$ & $1\mathrm e{-3}$ & 49.8 & 22.6 & 0/2\\
\midrule
\method & $2\mathrm e{-7}$ & 0 & 52.4 & 48.6 & 2/2\\
\method & $2\mathrm e{-7}$ & $1\mathrm e{-3}$ & 53.0 & 49.2 & 2/2\\
\method & $5\mathrm e{-7}$ & 0 & 51.6 & 47.0 & 2/2\\
\method & $5\mathrm e{-7}$ & $1\mathrm e{-3}$ & 52.2 & 47.8 & 2/2\\
\method & $1\mathrm e{-6}$ & 0 & 50.8 & 36.4 & 0/2\\
\method & $1\mathrm e{-6}$ & $1\mathrm e{-3}$ & 57.2 & 44.6 & 2/2\\
\bottomrule
\end{tabular}
\caption{Two seeds per LR--KL cell. Across 12 runs, PPO is stable in 3 and \method in 10.}
\label{tab:ppogrid}
\end{table}

\subsection{Frozen Qwen3-4B evaluation}
\begin{table}[H]
\centering
\scriptsize
\setlength{\tabcolsep}{3.2pt}
\begin{tabular}{lrrrrrrr}
\toprule
& & \multicolumn{3}{c}{Greedy pass@1} & \multicolumn{3}{c}{Majority@8}\\
Benchmark & $N$ & Base & GRPO & \method & Base & GRPO & \method\\
\midrule
AIME25 & 30 & 43.3 & 50.0 & 43.3 & 66.7 & 66.7 & 66.7\\
AIME2024 & 30 & 63.3 & 53.3 & 70.0 & 76.7 & 73.3 & 80.0\\
MATH500 & 500 & 67.2 & 66.6 & 69.8 & 79.8 & 79.2 & 81.0\\
GSM8K & 1319 & 91.7 & 93.1 & 94.1 & 93.5 & 94.3 & 94.5\\
LiveBench & 200 & 46.5 & 50.5 & 53.5 & 61.0 & 61.5 & 67.0\\
GPQA & 198 & 47.0 & 40.4 & 52.0 & 57.6 & 54.5 & 55.6\\
Olympiad & 674 & 48.5 & 48.2 & 49.4 & 53.0 & 51.6 & 52.1\\
\midrule
Macro & -- & 58.2 & 57.5 & \textbf{61.7} & 69.7 & 68.7 & \textbf{71.0}\\
\bottomrule
\end{tabular}
\caption{Frozen Qwen3-4B final-checkpoint point estimates.}
\label{tab:qwenfrozen}
\end{table}

\subsection{Frozen Llama-3.1-8B-Instruct evaluation}
\begin{table}[H]
\centering
\scriptsize
\setlength{\tabcolsep}{3.6pt}
\begin{tabular}{lrrrrrr}
\toprule
& \multicolumn{3}{c}{Greedy pass@1} & \multicolumn{3}{c}{Majority@8}\\
Benchmark & Base & GRPO & \method & Base & GRPO & \method\\
\midrule
AIME25 & 3.3 & 13.3 & 16.7 & 6.7 & 23.3 & 30.0\\
AIME2024 & 13.3 & 20.0 & 23.3 & 16.7 & 33.3 & 36.7\\
MATH500 & 53.4 & 64.2 & 69.6 & 64.4 & 76.6 & 80.6\\
GSM8K & 81.3 & 86.4 & 87.6 & 85.6 & 90.4 & 91.2\\
LiveBench & 22.5 & 35.0 & 39.0 & 29.0 & 44.5 & 49.0\\
GPQA & 26.8 & 32.8 & 40.4 & 32.3 & 39.4 & 47.0\\
Olympiad & 17.1 & 21.1 & 23.3 & 21.1 & 26.7 & 29.8\\
\midrule
Macro & 31.1 & 39.0 & \textbf{42.8} & 36.5 & 47.8 & \textbf{52.0}\\
\bottomrule
\end{tabular}
\caption{Frozen Llama-3.1-8B-Instruct final-checkpoint point estimates.}
\label{tab:llamafrozen}
\end{table}

\subsection{Stronger GRPO group control}
\begin{table}[H]
\centering
\small
\begin{tabular}{lrrrr}
\toprule
Method & Trajectories/update & ID Best & ID Final & Frozen greedy macro\\
\midrule
GRPO $4\!\times\!4$ & 16 & 56.4 & 53.8 & 57.5\\
GRPO $8\!\times\!8$ & 64 & 58.0 & 56.2 & 60.2\\
\method reference & 16 & 61.8 & 61.4 & 61.7\\
\bottomrule
\end{tabular}
\caption{Single-run stronger-group control. The \method row is the submitted-seed reference.}
\label{tab:stronggrpo}
\end{table}

%% file: sections/C_reproducibility.tex
\section{Reproducibility Details}
\label{app:reproducibility}

\subsection{Primary configuration}
\begin{table}[H]
\centering
\small
\begin{tabular}{lll}
\toprule
Item & GRPO & \method\\
\midrule
Backbone & Qwen3-4B & same\\
Transfer backbone & Llama-3.1-8B-Instruct & same\\
Maximum response & 16K & 16K\\
Batch / responses / minibatch & 4 / 4 / 4 & same\\
Actor learning rate & $10^{-6}$ & $10^{-6}$\\
Critic learning rate & -- & $10^{-5}$\\
KL coefficient & $10^{-3}$ & $2\times10^{-3}$\\
Policy clip & 0.20 & 0.20\\
Credit estimator & group-relative & $\kappa_t$ backup; $\lambda=0.95$\\
Gate map & -- & configured envelope $(0.1,0.9)$, $\tau=4$\\
Value clamp & -- & $[-1,1]$\\
Optimizer & AdamW $(0.9,0.999)$ & same\\
Weight decay / grad clip & 0 / 1 & same\\
Update epochs & 2 & 2\\
Sampling temperature / top-$p$ & 1.0 / 0.7 & same\\
Precision & BF16 & BF16\\
\bottomrule
\end{tabular}
\caption{Selected primary settings. Both methods were screened on the same actor-LR/KL grid.}
\label{tab:hyperparams}
\end{table}

\subsection{Phase controller}
The full performance experiments and all four factorial cells use the same closed-loop controller. It changes only after the predeclared stability trigger persists; escalation requires two consecutive trigger events and de-escalation requires five clear steps.

\begin{table}[H]
\centering
\small
\begin{tabular}{lrrrrr}
\toprule
Phase & KL & Clip & Actor LR & Value-loss clip & Critic LR mult.\\
\midrule
Stable & 0.0020 & 0.20 & $10^{-6}$ & 0.5 & 1.0\\
Warning & 0.0035 & 0.16 & $7\times10^{-7}$ & 0.4 & 0.9\\
Strong & 0.0045 & 0.14 & $5\times10^{-7}$ & 0.3 & 0.7\\
Hard-stop & 0.0070 & 0.09 & $2\times10^{-7}$ & 0.2 & 0.5\\
\bottomrule
\end{tabular}
\caption{Adaptive stability phases. The same controller is held fixed across the factorial, so it cannot induce the gate-by-critic interaction.}
\label{tab:controller}
\end{table}

The monitored signals are policy KL, gradient norm, clip fraction, entropy, and ID validation accuracy. Because validation participates in phase control, the 500-problem set is a development/held-out-ID set rather than a final test set. Frozen benchmarks are never used by the controller, checkpoint selection, or hyperparameter search.

\paragraph{Controller inputs and hysteresis.}
The controller maintains exponential moving averages with coefficient $0.4$ for policy clip fraction, PPO KL, KL loss, and gradient norm, and coefficient $0.2$ for entropy. It also reads raw response-clip ratio and ID development accuracy when evaluated. Entropy decline is measured against the oldest four values in an eight-step window. The target phase is the maximum severity triggered by the following rules:
\begin{itemize}
    \item entropy EMA below $8.0/7.5/6.5$ requests warning/strong/hard-stop;
    \item clip fraction at least $0.14/0.18/0.25$ requests warning/strong/hard-stop;
    \item PPO KL at least $0.08/0.12/0.22$ requests warning/strong/hard-stop;
    \item KL loss at least $2.85/3.20/4.00$ requests warning/strong/hard-stop;
    \item gradient norm at least $40/80$ requests warning/strong;
    \item development accuracy drop from the running best of at least $0.04/0.08/0.20$ requests warning/strong/hard-stop;
    \item an entropy drop of at least $1.5/3.0$, when clip fraction or PPO KL is already in warning territory, requests strong/hard-stop; and
    \item response-clip ratio at least $0.65$ requests hard-stop.
\end{itemize}
Escalation requires two consecutive steps with a higher target phase; de-escalation requires five consecutive clear steps. Fresh runs start in the stable phase. On fresh phase-0 runs, steps 25--40 linearly interpolate stable to warning knobs. If a screened launch KL differs from $0.002$, later phases preserve the same multipliers $1.75\times$, $2.25\times$, and $3.5\times$ relative to that launch value.

\paragraph{Additional phase-controlled quantities.}
The stable/warning/strong/hard-stop phases use advantage clips $5.0/4.0/3.5/2.5$, actor gradient clips $1.0/0.8/0.7/0.5$, actor epochs $2/2/2/1$, and critic gradient clips $1.0/0.8/0.7/0.5$. A reward schedule, applied in parallel rather than triggered by phase, ramps the format-penalty coefficient from $0.2$ to $1.0$ over steps 0--40; the length penalty is enabled at step 20 and ramps from $3\times10^{-5}$ to $8\times10^{-5}$ over steps 20--60. The PPO stress experiment disables critic warm-up and this controller; the fixed value-domain clamp remains enabled for the applicable value path.

\subsection{Value clamp control}
The clamp is a fixed output-domain constraint, not a method-specific adaptive intervention. In the most collapse-prone cell (Qwen3-4B, actor LR $10^{-6}$, KL 0), PPO and \method were each repeated three times without the clamp. The last out-of-range critic prediction occurred around step 37 for PPO and step 32 for \method on average. In the bounded grid, activation was therefore confined to comparable early critic transients and then became the identity. The stress-grid comparison applies the same bound to both methods.

\subsection{Attention extraction and memory}
The implementation uses final-layer GQA-group attention. Only the final layer exposes chunked attention; earlier layers retain fused attention kernels. Response-window-only buffers avoid prompt-heavy storage, online softmax/recomputation avoids writing dense attention to HBM, and prior-guided causal top-$K$ pooling uses $K=64$. The resulting critic adds fewer than $0.5\%$ trainable parameters and does not require a second same-scale Transformer or a second full optimizer state.

Exact GPU model/count, wall-clock time, peak allocated/reserved HBM, generated-token count, and throughput were not retained in the experiment record available for this version. They cannot be reconstructed reliably from public model specifications and are therefore not estimated. Consequently, the present paper does not make a measured wall-clock or peak-memory superiority claim.

\subsection{Statistical conventions}
Training intervals are two-sided 95\% Student-$t$ intervals over five independent runs. The same seed set is used across factorial cells, enabling seed-matched interaction intervals. Best is reported for completeness, but final is primary and last-20/AULC are secondary. Frozen results are currently point estimates from final checkpoints; they reflect finite benchmark sampling and checkpoint variance that is not quantified by the training-seed intervals.

\subsection{Data and result availability}
Editable CSV versions of every numerical table are included with the source archive. The public code release should additionally include the rollout configuration, evaluator scripts, checkpoint-selection rules, and commands needed to reproduce the training and frozen-evaluation tables.

%% file: sections/D_proofs.tex
\section{Proofs and Limiting Cases}
\label{app:proofs}

\subsection{Strict reduction}
If $\kappa_t=\gamma$ for all valid $t$, then \cref{eq:tkappa} becomes
\[
(\mathcal T_\kappa V)(h_t)=\E[r_t+\gamma V(h_{t+1})\mid h_t]
=(\mathcal T_\gamma V)(h_t).
\]
Substitution into \cref{eq:comp_td,eq:comp_gae} yields the standard TD residual and GAE recursion in \cref{eq:stdgae}. The outer objective \cref{eq:ppo_outer} is unchanged by construction, so the fixed-gate estimator path recovers the standard GAE/PPO path when the critic is also restored to the standard control.

\subsection{Contraction of the detached operator}
For bounded functions $V,W$ and any history $h_t$,
\begin{align*}
|\mathcal T_\kappa V(h_t)-\mathcal T_\kappa W(h_t)|
&=\left|\E\left[\kappa_t(V(h_{t+1})-W(h_{t+1}))\mid h_t\right]\right|\\
&\le \E\left[\kappa_t|V(h_{t+1})-W(h_{t+1})|\mid h_t\right]\\
&\le \kappamax\|V-W\|_\infty.
\end{align*}
Taking the supremum over $h_t$ proves the contraction. Banach's fixed-point theorem gives a unique bounded fixed point for a fixed policy and detached gate.

\subsection{Bounded trace}
Unrolling \cref{eq:comp_gae} gives \cref{eq:unrolled}. If $|\delta_t^\kappa|\le D$ and $\kappa_t\le\kappamax$, then
\begin{align*}
|A_t^{\kappa,\lambda}|
&\le\sum_{\ell=0}^{T-t}\lambda^\ell
\left(\prod_{j=0}^{\ell-1}\kappa_{t+j}\right)D\\
&\le D\sum_{\ell=0}^{\infty}(\lambda\kappamax)^\ell
=\frac{D}{1-\lambda\kappamax}.
\end{align*}

\subsection{Length normalization of attention concentration}
For a probability vector over $n_t$ valid historical positions,
\[
\frac{1}{n_t}\le H_t=\sum_i a_{ti}^2\le 1,
\]
where the lower bound follows from Cauchy--Schwarz and is achieved by the uniform vector, while the upper bound is achieved by a point mass. Therefore $n_tH_t\in[1,n_t]$, so
\[
0\le\frac{\log(n_tH_t)}{\log n_t}\le 1.
\]
At uniform attention, $n_tH_t=1$ and $c_t=0$; at a point mass, $n_tH_t=n_t$ and $c_t=1$. Because logarithm is monotone, the transformation preserves the ordering of $H_t$ at each $n_t$ while aligning the extremes across positions.

\subsection{Why the coefficient is not a post-hoc loss weight}
Multiplying a policy loss by an importance score changes the magnitude of the gradient at token $t$ but leaves the value target and propagation path unchanged. In \method, $\kappa_t$ enters both terms that define the estimator:
\[
\delta_t^\kappa=r_t+\kappa_tV_{t+1}-V_t,
\qquad
A_t^{\kappa,\lambda}=\delta_t^\kappa+\lambda\kappa_tA_{t+1}^{\kappa,\lambda}.
\]
It therefore changes the one-step bootstrap and every later residual's retention weight in \cref{eq:unrolled}. Its additional use in \cref{eq:value_mix} changes the state representation used to estimate the corresponding target. The policy loss consumes the resulting advantage without a direct multiplicative attention weight.

%% file: sections/E_limitations.tex
\section{Limitations, Open Questions, and Broader Impact}
\label{app:limitations}

\paragraph{Attention is a proxy, not causal attribution.}
Concentration can be distorted by attention sinks, positional biases, delimiter tokens, induction heads, or final-layer calibration. The method does not establish that highly attended tokens caused the outcome. A future version should compare layers and heads, mask known sinks, test intervention-based relevance, and use graph-valued rather than scalar signals.

\paragraph{Policy dependence and staleness.}
The gate is computed by the behavior policy and held fixed through multiple actor epochs. This is operationally simple and avoids differentiating the estimator through attention, but it creates staleness as the actor moves. Theoretical work should quantify the interaction among gate drift, PPO ratios, critic error, and clipping; empirical work should compare recomputation, fewer epochs, and explicit consistency regularization.

\paragraph{Choice of concentration statistic.}
Herfindahl concentration is bounded, monotone, inexpensive, and length-normalizable, but it is one point in a larger space. Shannon/R\'enyi entropy, top-$k$ coverage, head-specialist statistics, multi-layer aggregation, learned bounded maps, and attention-flow measures may be stronger. The present paper establishes that one trajectory-specific policy-internal statistic is useful, not that it is optimal.

\paragraph{Reward quality.}
Computation-conditioned transport can amplify a bad reward just as easily as a good one. Checker false negatives, formatting artifacts, or answer conventions may cause the algorithm to transport spurious evidence. Because the gate is correlated with the policy computation, reward artifacts that themselves dominate attention could be reinforced more strongly. Robust verifiers and audit sets remain essential.

\paragraph{Scope.}
The current evidence is mathematical reasoning on Qwen3-4B and Llama-3.1-8B-Instruct, with sparse terminal reward and responses up to 16K. Dialogue, tools, code, retrieval, multimodal generation, agents, dense rewards, and larger models are not experimentally established. The framework applies syntactically to these settings, but usefulness depends on whether the chosen internal statistic captures value-relevant computation.

\paragraph{Statistical and systems limitations.}
Training uncertainty is reported for the principal Qwen comparisons, but frozen evaluations and the $8\!\times\!8$ GRPO control are point estimates. Exact hardware, wall-clock, throughput, and peak-memory records are unavailable, so the current evidence does not support a measured systems-efficiency comparison.

\paragraph{Potential positive impact.}
More accurate credit transport may reduce rollout and training requirements, improve the viability of small actor-feature critics, and make long-horizon reasoning or agent training more sample efficient. The architecture-aware framing may also encourage modular combinations of credit estimators with better sampling, verifiers, and update geometry.

\paragraph{Potential risks.}
The method is a general optimizer for capable generative models and can improve downstream systems with both beneficial and harmful uses. Internal-signal optimization may also create new forms of reward hacking: a policy could learn attention or routing patterns that manipulate the estimator without improving task behavior. Detached gates and current controls do not rule out such co-adaptation over many training iterations. Future deployments should monitor gate distributions, test counterfactual stability, retain external behavioral evaluation, and avoid treating internal routing as a safety guarantee.